\documentclass[12pt]{article}
\usepackage{amsmath,epsfig,amssymb,amsfonts,amsthm,epsfig,verbatim,color,lscape,longtable,natbib,epstopdf,subfigure,graphicx,ltxtable,dcolumn,array,multirow,booktabs,adjustbox,url,mathrsfs,algorithm,algorithmic,appendix,adjustbox,rotating,tablefootnote}
\pdfoutput=1

\usepackage{setspace}
\usepackage{fullpage}
\usepackage{arydshln}

\newtheorem{Lem}{Lemma}
\newtheorem{Con}{Condition}

\newtheorem{Mth}{Theorem}

\newtheorem{Rem}{Remark}

\title{Transfer learning for tensor Gaussian graphical models}
\author{
	Mingyang Ren$^{1}$, Yaoming Zhen$^{2}$ and Junhui Wang$^{1}$ \\
	$^1$Department of Statistics \\
	The Chinese University of Hong Kong\\
	$^2$School of Data Science \\
	City University of Hong Kong \\
}
\date{}

\begin{document}
	
	\maketitle
	
	\onehalfspacing
	\begin{abstract}
		\vspace*{0.0em} \noindent 
		Tensor Gaussian graphical models (GGMs), interpreting conditional independence structures within tensor data, have important applications in numerous areas. Yet, the available tensor data in one single study is often limited due to high acquisition costs. Although relevant studies can provide additional data, it remains an open question how to pool such heterogeneous data. In this paper, we propose a transfer learning framework for tensor GGMs, which takes full advantage of informative auxiliary domains even when non-informative auxiliary domains are present, benefiting from the carefully designed data-adaptive weights. Our theoretical analysis  shows substantial improvement of estimation errors and variable selection consistency on the target domain under much relaxed conditions, by leveraging information from auxiliary domains. Extensive numerical experiments are conducted on both synthetic tensor graphs and a brain functional connectivity network data, which demonstrates the satisfactory performance of the proposed method.
		
	\end{abstract}
		
	\doublespacing
	\section{Introduction}
	
	The development of modern science facilitates collection of high-order tensor data in various research areas, ranging from molecular biology, neurophysiology, to signal processing. For examples, in cancer staging studies, multi-stage, multi-tissue, and multi-omics observations will be analyzed, which are organized as order-3 tensors \citep{krishnan2018integration}; in brain functional connectivity analysis, the functional magnetic resonance imaging (fMRI) data is also considered as an order-2 tensor, which includes blood oxygen level signals in different brain regions at different time points \citep{bellec2017neuro, zhang2019tensor}. 
	
	In light of the importance of tensor data in modern science, tensor data analysis has received increasing attention in recent years, such as supervised learning represented by tensor regression and classification \citep{zhou2013tensor, sun2017store, pan2018covariate} and unsupervised learning represented by tensor clustering and principal component analysis \citep{hopkins2015tensor, luo2022tensor}. In addition, the Gaussian graphical model (GGM) interpreting conditional independence structures within tensor data is also an essential topic but relatively understudied in literature. A straightforward approach for describing conditional independence in tensor data is to vectorize the tensor and fit multivariate GGMs \citep{friedman2008sparse, lam2009sparsistency, zhang2014sparse, liu2015fast}, which is considered, however, to largely ignore the tensor structure and require almost unrealistic estimation of a tremendous number of parameters \citep{he2014graphical}. For example, in the brain fMRI tensor data, if modeling the vectorized tensor with more than 200 time points and 116 widely studied brain regions of interest using multivariate GGMs, it requires estimation of more than 200 million parameters. More severely, simply vectorizing the tensor data may dilute our concern on the conditional independence between brain regions, corresponding to the functional brain connectivity, which is important for exploring the neurophysiological etiology. Tensor GGMs \citep{he2014graphical, lyu2019tensor} and related efficient algorithms \citep{min2022fast} are proposed in recent literature, and have been widely reported  their success. The models usually assume that the covariance matrix of the tensor data is separable, in the sense that it can be decomposed as the Kronecker product of multiple much smaller covariance matrices, each corresponding to one mode of the tensor data.  
	
	In many medical applications, high-dimensional and high-order tensor data are often extremely limited in one medical institution, due to the high acquisition costs and the rarity of certain diseases \citep{westin2002processing}. Fortunately, relevant data may be collected by other institutions, which may be helpful for the tasks studied at the target institution. Our motivation is to investigate the brain fMRI scans of  attention deficit hyperactivity disorder (ADHD) patients from various sites, in which the data in NeuroIMAGE site consists only 17 samples, but other sites can further provide more than ten times of relevant data. To pool these heterogeneous data from different sites, transfer learning is a promising solution with growing popularity, which aims at transferring the information from different auxiliary domains to help with the specific task on the target domain of interest \citep{pan2009survey}. 
	
	Transfer learning has been studied in many branches of machine learning, including image recognition \citep{gao2018deep}, natural language processing \citep{ruder2019transfer}, and drug discovery \citep{cai2020transfer}. More discussion on transfer learning can be found in \cite{zhuang2020comprehensive} and the references therein. Despite significant successes of transfer learning in algorithm developments and real-life applications, it is recognized that the existing studies on their statistical theory guarantees are still insufficient and is also gaining attention. Recently, \cite{cai2021transfer} proposes some minimax and adaptive transfer learning-based classifiers, \cite{bastani2021predicting} derives the estimation error bound of linear models in the single auxiliary domain case. \cite{li2022transfera} proposes the Trans-Lasso method under high-dimensional linear models with multiple auxiliary domains and establishes its minimax optimality. This transfer learning framework is extended to high-dimensional generalized linear models \citep{tian2022transfer}, federated learning \citep{li2021targeting}, and functional linear regression \citep{lin2022transfer}. However, transfer learning for unsupervised tasks, such as GGMs, is still in its infancy. It was only until very recently that \cite{li2022transfer} proposes a Trans-CLIME method for transfer learning on high-dimensional GGMs and it is subsequently extended to semiparametric graphical models \citep{he2022transfer}, but these approaches are still restricted to vector-value data.
	
	In this paper, we propose a transfer learning framework for tensor GGMs. It introduces a type of divergence matrix to measure the similarity between the target and auxiliary domains for each mode benefiting from the separability of the tensor covariance matrix, as well as some novel data-adaptive weights on the auxiliary domains based on the divergence matrices. The divergence matrix is first estimated based on a carefully designed regularized loss function by combining information from both target and auxiliary domains, and then the estimates of precision matrices can be better constructed based on the auxiliary domain and the well-estimated divergence matrices. The efficient algorithm and rigorous theoretical analysis of the proposed method are also conducted.
	
	This paper advances the current research on transfer learning in a number of ways. First, the proposed transfer learning method provides a more flexible modeling framework for high-order tensor GGMs, which also includes \cite{li2022transfer} as a special case. Second, to prevent the negative transfer phenomenon \citep{shu2019transferable}, data-adaptive weights for auxiliary domains are constructed to minimize the interference from the non-informative auxiliary domains. Third, the established theoretical analysis shows that the estimation error can be improved using the data-adaptive weights as long as there are at least one informative auxiliary domain that is close enough to the target domain. This is significantly different from the results in \cite{li2022transfer, he2022transfer}, which require all auxiliary domains to be informative for the improvement of error. Our theoretical analysis also demonstrates that transfer learning can help improve variable selection performance by weakening the regular minimum signal condition in literature \citep{lyu2019tensor}. Last but not least, the proposed method is applied to analyze the ADHD brain functional connectivity, which provides interesting neurophysiological insights in the pathogenesis.

	The rest of the paper is organized as follows. Section 2 introduces some necessary notations and brief backgrounds on tensor GGMs. Section 3 introduces the proposed transfer learning framework for the tensor GGMs and its implementing algorithm. The consistency of estimation and variable selection is established in Section 4. Numerical simulations and the application on ADHD brain fMRI data are conducted in Sections 5 and 6, respectively. Section 7 contains a brief discussion, and all technical details are provided in Supporting Information.

	\section{Preliminaries}
	In this section, we introduce necessary notations that will be used through out the paper and some brief backgrounds on tensor graphical model. 
	
	\subsection {Notations}
	
	Denote $\| \boldsymbol{u} \|_q$ as the  $l_q$-norm of a vector $\boldsymbol{u}$, for $q \geqslant 0$. For a matrix $\boldsymbol{A} = [A_{(i,j)}]_{1 \leqslant i,j \leqslant p}$, let $\boldsymbol{A}_{(j)}$ be its $j$-th column, $\| \boldsymbol{A} \|_{q, \infty} = \max_{1 \leqslant j \leqslant p} \| \boldsymbol{A}_{(j)} \|_q$, $\| \boldsymbol{A} \|_1 = \sum_{j=1}^{p} \| \boldsymbol{A}_{(j)} \|_1$, $\| \boldsymbol{A} \|_{\max} = \max_{1 \leqslant i,j \leqslant p} | A_{(i,j)} |$, $\| \boldsymbol{A} \|_{1, \mathrm{off}} = \sum_{1 \leqslant i \neq j \leqslant p} |A_{(i,j)}|$, and $\| \boldsymbol{A} \|_F$ be the Frobenius norm of $\boldsymbol{A}$. When $\boldsymbol{A}$ is symmetric, we further denote $\psi_{\min}(\boldsymbol{A})$ and $\psi_{\max}(\boldsymbol{A})$ as the smallest and largest eigenvalues of $\boldsymbol{A}$, respectively. A multidimensional array $\boldsymbol{\mathcal{X}} = (x_{i_1, \cdots, i_M}) \in \mathbb{R}^{p_1 \times \cdots \times p_M}$ is called a tensor of order-$M$. The vectorization of $\boldsymbol{\mathcal{X}}$ is defined by $\mathrm{vec}(\boldsymbol{\mathcal{X}}) \in \mathbb{R}^p$ with $p = \prod_{m=1}^M p_m$. The mode-$m$ matricization of $\boldsymbol{\mathcal{X}}$ is denoted by $\boldsymbol{\mathcal{X}}_{(m)} \in \mathbb{R}^{p_m \times (p/p_m)}$, which is obtained by arranging the mode-$m$ fibers of $\boldsymbol{\mathcal{X}}$ to be the columns of the resulting matrix. Herein, a mode-$m$ fiber of $\boldsymbol{\mathcal{X}}$  refers to a vector from $\boldsymbol{\mathcal{X}}$ by fixing all the indexes but the $m$-th mode.
	The mode-$m$ product between a tensor $\boldsymbol{\mathcal{X}}$ and a matrix $\boldsymbol{\Omega} \in \mathbb{R}^{d \times p_m}$ is defined as $\boldsymbol{\mathcal{X}} \times_{m} \boldsymbol{\Omega} \in \mathbb{R}^{p_1 \times \cdots p_{m-1} \times d \times p_{m+1} \times \cdots \times p_M}$, whose entry is defined as $(\boldsymbol{\mathcal{X}} \times_{m} \boldsymbol{\Omega})_{i_1, \cdots, i_{m-1}, j, i_{m+1} \cdots, i_M} = \sum_{i_m = 1}^{p_m} x_{i_1, \cdots, i_M} \boldsymbol{\Omega}_{j,i_m}$. In addition, for a list of matrices $\{ \boldsymbol{\Omega}_1, \cdots, \boldsymbol{\Omega}_M \}$ with $\boldsymbol{\Omega}_m \in \mathbb{R}^{d_m \times p_m}$, we define $ \boldsymbol{\mathcal{X}} \times \{ \boldsymbol{\Omega}_1, \cdots, \boldsymbol{\Omega}_M \} = \boldsymbol{\mathcal{X}} \times_1 \boldsymbol{\Omega}_1 \cdots \times_M \boldsymbol{\Omega}_M$. Similar to the matrix case, the Frobenius norm of $\boldsymbol{\mathcal{X}}$ is denoted as $\| \boldsymbol{\mathcal{X}} \|_F = (\sum_{i_1, \cdots, i_M} x_{i_1, \cdots, i_M}^2 )^{1/2}$. More detailed tensor algebra can be found in \cite{kolda2009tensor}.
	
	Finally, let $\text{card}(S)$ be the cardinality of a set $S$ and $[K] = \{ 1, \cdots, K \}$ be the $K$-set for any positive integer $K$. For sequences $a_n$ and $b_n$, define $a_n \lesssim b_n$ if there exists a positive constant $C$ such that $a_n \leqslant C b_n$, and  $ a_n\asymp b_n$ if $a_n \lesssim b_n$ and $b_n \lesssim a_n$. For two real numbers $a$ and $b$, define $a \wedge b = \min \{ a, b\}$ and $a \vee b = \max \{ a, b\}$. The superscript $^*$ of the parameter marks its true value.
	
	\subsection{Tensor GGMs}
	
	Suppose that an order-$M$ tensor $\boldsymbol{\mathcal{X}} = (x_{i_1, \cdots, i_M}) \in \mathbb{R}^{p_1 \times \cdots \times p_M}$ follows a zero-mean tensor normal distribution, denoted as $\boldsymbol{\mathcal{X}} \sim \mathrm{TN}(\boldsymbol{0}; \boldsymbol{\Sigma}_1, \cdots, \boldsymbol{\Sigma}_M)$, its probability density function is then defined as
	\begin{equation}\label{densi}
		\begin{aligned}
			p\left(\boldsymbol{\mathcal{X}} \mid \boldsymbol{\Sigma}_1, \ldots, \boldsymbol{\Sigma}_M \right) = (2 \pi)^{-p / 2}\left( \prod_{m=1}^M |\boldsymbol{\Sigma}_m |^{-p / (2p_{m})} \right)
			\exp \left(-\frac{1}{2} \left\| \boldsymbol{\mathcal{X}} \times \{\boldsymbol{\Sigma}^{-1 / 2} \} \right\|_F^2\right),
		\end{aligned}
	\end{equation}
	where $\boldsymbol{\Sigma}_m \in \mathbb{R}^{p_m \times p_m}$ is the mode-$m$ covariance matrix, and $\{\boldsymbol{\Sigma}^{-1 / 2} \} = \{ \boldsymbol{\Sigma}_1^{-1 / 2}, \cdots, \boldsymbol{\Sigma}_M^{-1 / 2} \}$. 
	Clearly, the tensor normal distribution extends the multivariate normal distribution \citep{tong2012multivariate} with $M=1$ or matrix normal distribution \citep{matsuda2022estimation} with $M=2$ to a tensor random variable with general order-$M$. It can be shown that $\boldsymbol{\mathcal{X}} \sim \mathrm{TN}(\boldsymbol{0}; \boldsymbol{\Sigma}_1, \cdots, \boldsymbol{\Sigma}_M)$ if and only if $\mathrm{vec}(\boldsymbol{\mathcal{X}}) \sim N(\mathrm{vec}(\boldsymbol{0}) ; \boldsymbol{\Sigma}_M \otimes \cdots \otimes \boldsymbol{\Sigma}_1)$, where $\otimes$ stands for the Kronecker product. To account for the identifiability issue, we follow the common treatment \citep{lyu2019tensor} and require that $\| \boldsymbol{\Omega}_m^* \|_F = 1$ for $m \in [M]$, where $\boldsymbol{\Omega}_m^* = (\boldsymbol{\Sigma}^*_m)^{-1}$ is the precision matrix in the $m$-th mode.
	
	We consider sparse estimation of $\{ \boldsymbol{\Omega}_m \}_{m=1}^{M}$ to characterize the conditional independence relation among the features of any given mode of $\boldsymbol{\mathcal{X}}$. Specifically, let  $\boldsymbol{\mathcal{X}}^{(m)}_{(i)} \in \mathbb{R}^{p_1 \times \ldots p_{m-1} \times p_{m+1} \times \ldots \times p_{M}}$ denote the $i$-th sub-tensor extracted from $\boldsymbol{\mathcal{X}}$ by fixing the index in the $m$-th mode as $i$, then $[\boldsymbol{\Omega}_m]_{(i,i^{\prime})} = 0$ if and only if $\boldsymbol{\mathcal{X}}^{(m)}_{(i)}$ is independent of $\boldsymbol{\mathcal{X}}^{(m)}_{(i^{\prime})}$ given all other $\boldsymbol{\mathcal{X}}^{(m)}_{(j)}$ with $j \ne i, i^{\prime}$. For example, in an order-3 tendor $\boldsymbol{\mathcal{X}}$, $x_{i_1, i_2, i_3}$ denotes the activation level at region $i_1$ of subject $i_2$ in the $i_3$-th fMRI scan over the lateral prefrontal cortex, $[\boldsymbol{\Omega}_1]_{(i_1, i_1^\prime)}$ indicates the regularity strength of regions $i_1$ and $i_1^\prime$ given the activation levels of all other regions of interests across different subjects and scans, and the activation levels of region $i_1$ and $i_1^\prime$ are conditional independent if and only if $[\boldsymbol{\Omega}_1]_{(i_1, i_1^\prime)} = 0$.
	
	Estimation of $\boldsymbol{\Omega}_m$ amounts to maximizing the likelihood function of $\{ \boldsymbol{\mathcal{X}}_i \}_{i=1}^{n}$ that are independently sampled from (\ref{densi}), which is block multi-convex \citep{lyu2019tensor} with respect to $\{\boldsymbol{\Omega_m}\}_{m=1}^M$. 	Leveraging the multi-convex property, \cite{lyu2019tensor} proposed to alternatively update one precision matrix with others fixed. Specifically, one can minimize
	\begin{equation}\label{glassoten}
		\begin{aligned}
			\ell ( \boldsymbol{\Omega}_m ) = -\frac{1}{p_m} \log [\det (\boldsymbol{\Omega}_m)] + \frac{1}{p_m} \operatorname{tr} (\boldsymbol{S}_m \boldsymbol{\Omega}_m ) + \lambda_m\left\|\boldsymbol{\Omega}_m\right\|_{1, \mathrm{off}},
		\end{aligned}
	\end{equation}
	where  $\boldsymbol{S}_m = \frac{p_m}{n p} \sum_{i=1}^{n} \boldsymbol{V}_{i(m)} \boldsymbol{V}_{i(m)}^{\top}$, $\boldsymbol{V}_{i(m)} = [\boldsymbol{\mathcal{X}}_{i}]_{(m)} ( \boldsymbol{\Omega}_M^{1/2} \otimes \cdots \otimes \boldsymbol{\Omega}_{m+1}^{1/2} \otimes \boldsymbol{\Omega}_{m-1}^{1/2} \otimes \cdots \otimes\boldsymbol{\Omega}_1^{1/2} )$, and $\det(\boldsymbol{\Omega}_m)$ is the determinant of $\boldsymbol{\Omega}_m$. This optimization task can be efficiently solved via the graphical lasso algorithm \citep{friedman2008sparse}, and the obtained estimates of $\boldsymbol{\Omega}_m$'s enjoy the asymptotic consistency following standard treatment of penalized maximum likelihood estimation \citep{lyu2019tensor}. Yet, the applicability of such consistency results requires a sufficiently large sample size, which is usually not realistic in practice. To this end, we propose a transfer learning method to leverage information from auxiliary domains so as to enhance the learning performance in the target domain.

	\section{Proposed method}
	Suppose that besides observations $\{ \boldsymbol{\mathcal{X}}_i\}_{i=1}^n$ from the target domain, observations $\{ \boldsymbol{\mathcal{X}}_i^{(k)}\}_{i=1}^{n_k}$; $k \in [K]$ from some auxiliary domains are also available. For example, in the ADHD brain functional network dataset, $\{ \boldsymbol{\mathcal{X}}_i\}_{i=1}^n$ are the dynamic activation levels of many brain regions of interests collected from some fMRI scans at one neuroscience institute, and $\{ \boldsymbol{\mathcal{X}}_i^{(k)}\}_{i=1}^{n_k}$ are collected from $K=6$ other neuroscience institutes for better data analysis in the target institute. That is, $ \boldsymbol{\mathcal{X}}_i$'s are independently generated from $\mathrm{TN}(\boldsymbol{0}; \boldsymbol{\Sigma}_1, \cdots, \boldsymbol{\Sigma}_M)$ and $\boldsymbol{\mathcal{X}}_i^{(k)}$'s are independently generated from $ \mathrm{TN}(\boldsymbol{0}; \boldsymbol{\Sigma}_1^{(k)}, \cdots, \boldsymbol{\Sigma}_M^{(k)})$ with $\Sigma_m \in \mathbb{R}^{p_m \times p_m}$ and $\boldsymbol{\Sigma}_m^{(k)} \in \mathbb{R}^{p_m \times p_m}$.
	Particularly, we are interested in estimating the precision matrix  $\boldsymbol{\Omega}_m = (\boldsymbol{\Sigma}_m)^{-1}$ in the target domain for $m\in [M]$ via transfer learning on the tensor GGMs.

	\subsection{Divergence matrix}\label{diver}
	The key of transfer learning is to construct a similarity measure between parameters of interest in the auxiliary and target domains. Particularly, let $TN_{\Sigma^{(k)}}$ and $TN_{\Sigma}$ denote $\mathrm{TN}(\boldsymbol{0}; \boldsymbol{\Sigma}_1^{(k)}, \cdots, \boldsymbol{\Sigma}_M^{(k)})$ and $\mathrm{TN}(\boldsymbol{0}; \boldsymbol{\Sigma}_1, \cdots, \boldsymbol{\Sigma}_M)$ for short, and
	we consider the Kullback–Leibler (KL) divergence between $TN_{\Sigma^{(k)}}$ and $TN_{\Sigma}$,
	\begin{equation}\nonumber
		\begin{aligned}
			& KL (TN_{\Sigma^{(k)}} || TN_{\Sigma}) = - \sum_{m=1}^{M}\frac{p}{2p_m} \log [ \det ( \boldsymbol{\Omega}_m \boldsymbol{\Sigma}_{m}^{(k)} ) ]\\
			& \hspace{21ex} + \frac{1}{2} \left\{ \mathbb{E} \left( \| \boldsymbol{\mathcal{X}}^{(k)} \times \{\boldsymbol{\Omega}^{1 / 2} \} \|_F^2 \right) - \mathbb{E} \left( \| \boldsymbol{\mathcal{X}}^{(k)} \times \{ (\boldsymbol{\Omega}^{(k)})^{1 / 2} \} \|_F^2 \right) \right\},
		\end{aligned}
	\end{equation}
	where $\{\boldsymbol{\Omega}^{1 / 2} \} = \{ \boldsymbol{\Omega}_1^{1 / 2}, \cdots, \boldsymbol{\Omega}_M^{1 / 2} \}$ and $\{ (\boldsymbol{\Omega}^{(k)})^{1 / 2} \} = \{ (\boldsymbol{\Omega}^{(k)}_1)^{1 / 2}, \cdots, (\boldsymbol{\Omega}^{(k)}_M)^{1 / 2} \}$. 
	
	Define the {\it divergence matrix} as 
	$\boldsymbol{\Delta}_{m}^{(k)} = \boldsymbol{\Omega}_m \boldsymbol{\Sigma}_m^{(k)}  - \boldsymbol{I}_{p_m}$, where
	$\boldsymbol{I}_{p_m}$ is the $p_m$-dimensional identity matrix. 
	Clearly, it gets closer to $\boldsymbol{0}$ when $\boldsymbol{\Sigma}_m^{(k)}$ gets closer to $\boldsymbol{\Sigma}_m$, and thus it provides a natural measure of the similarity between $\boldsymbol{\Sigma}_m^{(k)}$ and $\boldsymbol{\Sigma}_m$. More interestingly, if $\boldsymbol{\Omega}_{m^{\prime}} = \boldsymbol{\Omega}_{m^{\prime}}^{(k)}$ for all $m^{\prime} \neq m$, it follows that
	$$
	KL (TN_{\Sigma^{(k)}} || TN_{\Sigma}) = - \frac{p}{2 p_{m}} \log [\det(\boldsymbol{\Delta}_{m}^{(k)} + \boldsymbol{I}_{p_{m}})] + \frac{p}{2 p_{m}} \operatorname{tr} [\boldsymbol{\Delta}_{m}^{(k)}],
	$$ 
	which is solely parametrized by $\boldsymbol{\Delta}_{m}^{(k)}$. 
	
	To leverage information of all auxiliary domains, we consider the weighted average of the covariance and divergence matrices as follows,
	\begin{equation}\nonumber
		\begin{aligned}
			& \boldsymbol{\Sigma}_m^{\mathcal{A}} = \sum_{k=1}^{K} \alpha_k \boldsymbol{\Sigma}_m^{(k)} \text{ and }
			\boldsymbol{\Delta}_{m} = \sum_{k=1}^{K} \alpha_k \boldsymbol{\Delta}_m^{(k)}, \text{ with } \sum_{k=1}^{K} \alpha_k = 1,
		\end{aligned}
	\end{equation}
	where the choice of weights $\{\alpha_k\}_{k=1}^{K}$ shall depend on the contribution of each auxiliary domain and will be discussed in details in Section \ref{weights}. Also, it holds true that $\boldsymbol{\Omega}_m \boldsymbol{\Sigma}_m^{\mathcal{A}} - \boldsymbol{\Delta}_{m} - \boldsymbol{I}_{p_m} = \boldsymbol{0}$.

	\subsection{Separable transfer estimation}
	For each $m \in [M]$, we first estimate $\boldsymbol{\Delta}_{m}$ via samples from both the auxiliary and target domains, and then estimate $\boldsymbol{\Omega}_{m}$ by leveraging only the auxiliary samples. Accordingly, we design two specific loss functions for $\boldsymbol{\Delta}_{m}$ and $\boldsymbol{\Omega}_{m}$ as
	\begin{equation}\nonumber
		\begin{aligned}
			& \mathcal{L}_{\Delta} (\boldsymbol{\Delta}_{m}; \{ \boldsymbol{\Sigma}_{m}^{(k)} \}_{k=1}^{K}, \boldsymbol{\Omega}_m ) = \frac{1}{2} \operatorname{tr} \{  \boldsymbol{\Delta}_{m}^{\top} \boldsymbol{\Delta}_{m} \} -\operatorname{tr} \{  \left( \boldsymbol{\Omega}_m \boldsymbol{\Sigma}_m^{\mathcal{A}} - \boldsymbol{I}_{p_m} \right) ^{\top} \boldsymbol{\Delta}_{m} \}, \\
			& \mathcal{L}_{\Omega} ( \boldsymbol{\Omega}_{m}; \{ \boldsymbol{\Sigma}_{m}^{(k)} \}_{k=1}^{K}, \boldsymbol{\Delta}_{m} ) = \frac{1}{2} \operatorname{tr} \{ \boldsymbol{\Omega}_{m}^{\top} \boldsymbol{\Sigma}_m^{\mathcal{A}} \boldsymbol{\Omega}_{m} \} -\operatorname{tr} \{ ( \boldsymbol{\Delta}_{m}^{\top}+ \boldsymbol{I}_{p_m} ) \boldsymbol{\Omega}_{m} \},
		\end{aligned}
	\end{equation}
	where $\boldsymbol{\Sigma}_m^{\mathcal{A}} = \sum_{k=1}^{K} \alpha_k \boldsymbol{\Sigma}_m^{(k)}$ for any $\{ \alpha_k \}_{k=1}^{K}$ satisfying $\sum_{k=1}^{K} \alpha_k = 1$. The two loss functions are expressed as the difference of two trace operators, which share similar spirit with the D-trace loss \citep{zhang2014sparse}. 
	\begin{Lem}\label{D-trace}
		Both loss functions $\mathcal{L}_{\Delta} (\boldsymbol{\Delta}_{m}; \{ \boldsymbol{\Sigma}_{m}^{(k)} \}_{k=1}^{K}, \boldsymbol{\Omega}_m )$ and $\mathcal{L}_{\Omega} ( \boldsymbol{\Omega}_{m}; \{ \boldsymbol{\Sigma}_{m}^{(k)} \}_{k=1}^{K}, \boldsymbol{\Delta}_{m} )$ are convex with respect to $\boldsymbol{\Delta}_{m}$ and $\boldsymbol{\Omega}_{m}$, respectively. Furthermore, $\boldsymbol{\Delta}_{m}^*$ and $\boldsymbol{\Omega}_{m}^*$ are unique minimizers of $\mathcal{L}_{\Delta} (\boldsymbol{\Delta}_{m}; \{ \boldsymbol{\Sigma}_{m}^{(k)*} \}_{k=1}^{K}, \boldsymbol{\Omega}_m^* )$ and $\mathcal{L}_{\Omega} ( \boldsymbol{\Omega}_{m}; \{ \boldsymbol{\Sigma}_{m}^{(k)*} \}_{k=1}^{K}, \boldsymbol{\Delta}_{m}^* )$, respectively. 
	\end{Lem}
	By Lemma \ref{D-trace}, the two empirical loss functions are suitable to get accurate estimation of $\boldsymbol{\Delta}_{m}^*$ and $\boldsymbol{\Omega}_{m}^*$.
	Furthermore, both empirical losses can be equipped with various regularization terms if additional structures are desired.

	In view of the above discussion, for each mode, a multi-step method can be proposed to realize the transfer learning of tensor graphical models.
	
	{\it Step 1}. Initialization. Estimate $\{ \widehat{\boldsymbol{\Omega}}^{(0)}_m \}_{m=1}^{M}$ based on target samples $\{ \boldsymbol{\mathcal{X}}_i \}_{i=1}^{n} $, and $\{ \widehat{\boldsymbol{\Omega}}_{m}^{(k)} \}_{m=1}^{M}$ based on auxiliary samples $\{ \boldsymbol{\mathcal{X}}_i^{(k)}\}_{i=1}^{n_k}$, for $k \in [K]$, using the separable estimation approach \citep{lyu2019tensor}. Then, define
	\begin{equation}\nonumber
		\begin{aligned}
			& \widehat{\boldsymbol{\Sigma}}_m^{\mathcal{A}} = \sum_{k=1}^{K} \alpha_k \widehat{\boldsymbol{\Sigma}}_m^{(k)}, \ \ \text{ where } \widehat{\boldsymbol{\Sigma}}_m^{(k)} =  \frac{p_m}{n_k p} \sum_{i=1}^{n_k}\widehat{\boldsymbol{V}}_{i,m}^{(k)} \widehat{\boldsymbol{V}}_{i,m}^{(k) \top}, \\
			& \widehat{\boldsymbol{V}}_{i,m}^{(k)} = [ \boldsymbol{\mathcal{X}}_{i}^{(k)}]_{(m)} \left[  ( \widehat{\boldsymbol{\Omega}}_M^{(k)} )^{1/2} \otimes \cdots \otimes ( \widehat{\boldsymbol{\Omega}}_{m+1}^{(k)} )^{1/2} \otimes (\widehat{\boldsymbol{\Omega}}_{m-1}^{(k)})^{1/2} \otimes \cdots \otimes (\widehat{\boldsymbol{\Omega}}_{1}^{(k)})^{1/2} \right].
		\end{aligned}
	\end{equation}
	
	{\it Step 2}. For each $m \in [M]$, perform the following two estimation steps separately.
	
	(a). Estimate the divergence matrix of mode-$m$,
	\begin{equation}\label{delta_obj}
		\begin{aligned}
			& \widehat{\boldsymbol{\Delta}}_{m} = \arg \min \mathcal{Q}_{1} (\boldsymbol{\Delta}_m ),
		\end{aligned}
	\end{equation}
	where $\mathcal{Q}_{1} (\boldsymbol{\Delta}_{m} ) = \frac{1}{2} \operatorname{tr} \{ \boldsymbol{\Delta}_{m}^{\top} \boldsymbol{\Delta}_{m} \}-\operatorname{tr} \left\{ ( \widehat{\boldsymbol{\Omega}}_m^{(0)} \widehat{\boldsymbol{\Sigma}}_m^{\mathcal{A}} - \boldsymbol{I}_{p_m} ) ^{\top} \boldsymbol{\Delta}_{m} \right\}+ \lambda_{1m} \| \boldsymbol{\Delta}_{m} \|_1$.
	
	(b). Estimate the precision matrix of mode-$m$,
	\begin{equation}\label{theta_obj}
		\begin{aligned}
			& \widehat{\boldsymbol{\Omega}}_{m} = \arg \min \mathcal{Q}_{2} ( \boldsymbol{\Omega}_{m} ),
		\end{aligned}
	\end{equation}
	where $\mathcal{Q}_{2} ( \boldsymbol{\Omega}_{m}) = \frac{1}{2} \operatorname{tr} \{ \boldsymbol{\Omega}_{m}^{\top} \widehat{\boldsymbol{\Sigma}}_m^{\mathcal{A}} \boldsymbol{\Omega}_{m} \} -\operatorname{tr} \{ ( \widehat{\boldsymbol{\Delta}}_{m}^{\top}+ \boldsymbol{I}_{p_m} ) \boldsymbol{\Omega}_{m} \} + \lambda_{2m} \| \boldsymbol{\Omega}_{m} \|_{1, \mathrm{off}}$.
	
	In Step 2(a), $\widehat{\boldsymbol{\Delta}}_m$ can be considered as an adaptive thresholding of a naive estimate, $\widehat{\boldsymbol{\Omega}}_m^{(0)} \widehat{\boldsymbol{\Sigma}}_m^{\mathcal{A}} - \boldsymbol{I}_{p_m}$, which is inspired by the definition of $\widehat{\boldsymbol{\Delta}}_m$. If the difference between the target and auxiliary domains in mode-$m$ precision matrices are small enough, some elements of $\widehat{\boldsymbol{\Delta}}_m$ can shrink to zero with appropriate $\lambda_{1m}$. The thresholding can improve the estimation of $\widehat{\boldsymbol{\Delta}}_m$ with the help of the auxiliary samples.
	Correspondingly, $\boldsymbol{\Omega}_{m}$ can also be better estimated via $\widehat{\boldsymbol{\Delta}}_m$ by leveraging only the auxiliary samples in Step 2(b).
	
	Moreover, the similarity between the target and auxiliary domains may be weak in some scenarios, so that the learning performance in the target domain may be deteriorated due to information transfer, which is so-called ``negative transfer" \citep{shu2019transferable}. One practical solution is to further perform a model selection step following \cite{li2022transfer}, which guarantees that transfer learning is no less effective than using only the target domain. To this end, the data from the target domain can be randomly split into two folds $\mathcal{N}$ and $\mathcal{N}^C$, satisfying $\mathcal{N} \cup \mathcal{N}^C = \{ 1, \cdots, n \}$ and $\text{card}(\mathcal{N}) = cn$, for some constant $0 < c <1$. The value of $c$ is not sensitive \citep{li2022transfer}, and we set $c=0.6$ in all numerical experiments. The subjects in $\mathcal{N}$ are used to construct the initialization of the separable transfer estimation in Step 1. The selection step is performed based on subjects in $\mathcal{N}^C$. Specifically, based on $\{ \widetilde{\boldsymbol{\Omega}}_{m}^{(0)} \}_{m=1}^{M}$ estimated using subjects in $\mathcal{N}^C$, for $j = 1, \cdots, p_m$, define $\widetilde{\boldsymbol{\Sigma}}_m =  \frac{p_m}{ (1-c)n p} \sum_{i \in \mathcal{N}^C} \widetilde{\boldsymbol{V}}_{i,m} \widetilde{\boldsymbol{V}}_{i,m}^{\top}$, $\widetilde{\boldsymbol{V}}_{i,m} = [\boldsymbol{\mathcal{X}}_{i}]_{(m)
	} \left[  ( \widetilde{\boldsymbol{\Omega}}_M^{(0)} )^{1/2} \otimes \cdots \otimes ( \widetilde{\boldsymbol{\Omega}}_{m+1}^{(0)} )^{1/2} \otimes (\widetilde{\boldsymbol{\Omega}}_{m-1}^{(0)} )^{1/2} \otimes \cdots \otimes (\widetilde{\boldsymbol{\Omega}}_{1}^{(0)} )^{1/2} \right]$, and
	\begin{equation}\nonumber
		\begin{aligned}
			& \widehat{w}_{m,j} = \underset{ w \in \{ (0,1)^{\top}, (1,0)^{\top} \} }{\arg\min} \|  \widetilde{\boldsymbol{\Sigma}}_m ( \widehat{\boldsymbol{\Omega}}^{(0)}_{m(j)}, \widehat{\boldsymbol{\Omega}}_{m(j)} ) w -  \boldsymbol{I}_{p_m(j)} \|_2^2,
		\end{aligned}
	\end{equation}
	where $\widehat{\boldsymbol{\Omega}}^{(0)}_{m(j)}$, $\widehat{\boldsymbol{\Omega}}_{m(j)}$, and $\boldsymbol{I}_{p_m(j)}$ are the $j$-th columns of $\widehat{\boldsymbol{\Omega}}^{(0)}_{m}$, $\widehat{\boldsymbol{\Omega}}_{m}$, and $\boldsymbol{I}_{p_m}$, respectively. 
	Then the final estimate becomes
	\begin{equation}\label{omega_obj}
		\widehat{\boldsymbol{\Omega}}_{m(j)}^{(f)} = (\widehat{\boldsymbol{\Omega}}^{(0)}_{m(j)}, \widehat{\boldsymbol{\Omega}}_{m(j)}) \widehat{w}_{m(j)}.
	\end{equation}
	The selection step realizes a model selection between the $\widehat{\boldsymbol{\Omega}}^{(0)}_{m(j)}$ and $\widehat{\boldsymbol{\Omega}}_{m(j)}$, which yields satisfactory theoretical and numerical performance \citep{li2022transfer}. Note that $\widehat{\boldsymbol{\Omega}}_{m}^{(f)}$ is not symmetric in general, and $(\widehat{\boldsymbol{\Omega}}_{m}^{(f)} + [\widehat{\boldsymbol{\Omega}}_{m}^{(f)}]^{\top}) / 2$ can be used as a symmetric estimate. Furthermore, it can be theoretically guaranteed that the final estimate is positive definite \citep{liu2015fast,li2022transfer}.

	\subsection{Construction of weights}\label{weights}
	
	A natural choice of the weights is to set 
	\begin{equation}\label{na-weight}
		\widehat{\boldsymbol{\Sigma}}_m^{\mathcal{A}} = \sum_{k=1}^{K} \alpha_k \widehat{\boldsymbol{\Sigma}}_m^{(k)}, \ \mbox{with} \ \alpha_k = n_k / N \ \mbox{and} \  N = \sum_{k=1}^{K} n_k,
	\end{equation}
	following from the fact that the auxiliary domain with larger sample size shall be more important. Yet, it does not take into account the similarities between the target and auxiliary domains. If there are some large non-informative auxiliary domains, although the final model selection step can guarantee that transfer learning is no less effective than using the target domain only, it may also offset the potential improvement benefiting from the informative auxiliary domains with positive impact.
	
	To address this challenge, we further design some data-adaptive weights for auxiliary covariance matrices, in which weights are constructed combining both sample sizes and the estimated differences between the target and auxiliary domains. Particularly, we set
	\begin{equation}\label{adap-weight}
		\begin{aligned}
			& \widehat{\boldsymbol{\Sigma}}_m^{\mathcal{A}} = \sum_{k=1}^{K} \alpha_k \widehat{\boldsymbol{\Sigma}}_m^{(k)}, \text{ with } \alpha_k = \frac{n_k / \widehat{h}_k}{\sum_{k=1}^{K} (n_k / \widehat{h}_k)},
		\end{aligned}
	\end{equation}
	where $\widehat{h}_k = \max_{m \in [M]} \| \widehat{\boldsymbol{\Delta}}_{m}^{(k)} \|_{1, \infty}$ and  $\widehat{\boldsymbol{\Delta}}_{m}^{(k)} = \widehat{\boldsymbol{\Omega}}_m^{(0)} \widehat{\boldsymbol{\Sigma}}_m^{(k)}  - \boldsymbol{I}_{p_m}$. Clearly, for auxiliary domains with similar sample size, the weight for the one with smaller difference from the target domain is larger. Here we note that the type of norm for measuring similarity is not critical, and the specified $L_1$-norm is only for keeping with the form on theoretical analysis and may be replaced by other norms with slight modification. It is also interesting to note that even with such data-adaptive weights, the model selection step in (\ref{omega_obj}) is still necessary to safeguard the extreme case where all the auxiliary domains are non-informative.

	\subsection{Computing algorithm}
	
	For Step 2(a), define $\widehat{\boldsymbol{B}}_m = \widehat{\boldsymbol{\Omega}}_m^{(0)} \widehat{\boldsymbol{\Sigma}}_m^{\mathcal{A}} - \boldsymbol{I}_{p_m}$ for each $m \in [M]$, and then (\ref{delta_obj}) can be rewritten as
	\begin{equation}\nonumber
		\begin{aligned}
			\mathcal{Q}_{1} (\boldsymbol{\Delta}_{m} ) 
			& = \frac{1}{2} \sum_{1 \leqslant i,j \leqslant p_m} [\boldsymbol{\Delta}_{m}]_{(i,j)}^2 - \sum_{1 \leqslant i,j \leqslant p_m} [\widehat{\boldsymbol{B}}_{m}]_{(i,j)} [\boldsymbol{\Delta}_{m}]_{(i,j)} + \lambda_{1m} \sum_{1 \leqslant i,j \leqslant p_m} | [\boldsymbol{\Delta}_{m}]_{(i,j)} |,
		\end{aligned}
	\end{equation}
	where $[\boldsymbol{\Delta}_{m}]_{(i,j)}$ and $[\widehat{\boldsymbol{B}}_{m}]_{(i,j)}$ are the $(i,j)$ entries of $\boldsymbol{\Delta}_{m}$ and $\widehat{\boldsymbol{B}}_m$, respectively. It can be separated into $p^2_m$ lasso-type optimizations; that is, for any $i$ and $j$, 
	$$
	[\widehat{\boldsymbol{\Delta}}_{m}]_{(i,j)} = \arg \min_{\Delta} \left\{  \frac{1}{2} (\Delta - [\widehat{\boldsymbol{B}}_{m}]_{(i,j)} )^2 + \lambda_{1m} | \Delta | \right\} = \mathcal{T}( [\widehat{\boldsymbol{B}}_{m}]_{(i,j)}, \lambda_{1m}),
	$$
	where $\mathcal{T}(z, \lambda) = \mathrm{sign}(z)  \max (0, |z| - \lambda)$.

	For Step 2(b), note that (\ref{theta_obj}) can be rewritten as
	\begin{equation}\nonumber
		\begin{aligned}
			\mathcal{Q}_{2} ( \boldsymbol{\Omega}_{m})
			& = \sum_{1 \leqslant j \leqslant p_m} \left\{ \frac{1}{2} \boldsymbol{\Omega}_{m(j)}^{\top} \widehat{\boldsymbol{\Sigma}}_m^{\mathcal{A}} \boldsymbol{\Omega}_{m(j)} - \boldsymbol{\Omega}_{m(j)}^{\top}( \widehat{\boldsymbol{\Delta}}_{m(j)}+ \boldsymbol{I}_{p_m(j)} )  + \lambda_{2m} \| \boldsymbol{\Omega}_{m(j)} \|_1 - \lambda_{2m} | [\boldsymbol{\Omega}_{m}]_{(j,j)} | \right\},
		\end{aligned}
	\end{equation}
	where $\boldsymbol{\Omega}_{m(j)}$ and $\boldsymbol{I}_{p_m(j)}$ are the $j$-th columns of $\boldsymbol{\Omega}_{m}$ and $\boldsymbol{I}_{p_m}$, respectively. It can be separated into $p_m$ optimizations; that is, for any $j$, 
	\begin{equation}\label{thetaj}
		\begin{aligned}
			\widehat{\boldsymbol{\Omega}}_{m(j)} = \arg \min_{\boldsymbol{\theta}} \left\{  \frac{1}{2} \boldsymbol{\theta}^{\top} \widehat{\boldsymbol{\Sigma}}_m^{\mathcal{A}} \boldsymbol{\theta} - \boldsymbol{\theta}^{\top}( \widehat{\boldsymbol{\Delta}}_{m(j)}+ \boldsymbol{I}_{p_m(j)} )  + \lambda_{2m} \| \boldsymbol{\theta}_{(-j)} \|_1  \right\},
		\end{aligned}
	\end{equation}
	where $\boldsymbol{\theta}_{(-j)}$ is the sub-vector of $\boldsymbol{\theta}$ with the $j$-th component removed.
	
	For the optimization of (\ref{thetaj}), we adopt the coordinate descent algorithm. Particularly, at iteration $t + 1$, the updating formula of $\theta_i$, $i$-th component of $\boldsymbol{\theta}$, with other components $\{ \theta_{i^{\prime}}^{(t+1)}, i^{\prime} < i; \  \theta_{i^{\prime}}^{(t)}, i^{\prime} > i \}$ fixed, are
	\begin{align}
		\theta^{(t+1)}_i = [\widehat{\boldsymbol{\Sigma}}_m^{\mathcal{A}}]_{(i,i)}^{-1} \mathcal{T}(\xi^{(t)}, \lambda_{2m} I(i \ne j)), \text{ for } i = 1, \cdots, p_m, \nonumber
	\end{align}
	where $\xi^{(t)} = [\widehat{\boldsymbol{\Delta}}_{m} + \boldsymbol{I}_{p_m}]_{(i,j)} - \sum_{i^{\prime} < i} \theta_{i^{\prime}}^{(t+1)} [\widehat{\boldsymbol{\Sigma}}_m^{\mathcal{A}}]_{(i,i^{\prime} )} - \sum_{i^{\prime} > i} \theta_{i^{\prime}}^{(t)} [\widehat{\boldsymbol{\Sigma}}_m^{\mathcal{A}}]_{(i,i^{\prime} )}$.
	
	As computational remarks, explicit solution can be derived in each step of the algorithm, which makes it very efficient. The initial values of $\boldsymbol{\theta}$ are set as $\widehat{\boldsymbol{\Omega}}_{m(j)}^{(0)}$. Note that these developments are specifically for the Lasso penalty, and optimization with other penalties may require minor modifications. 
	Convergence properties of the algorithm can be guaranteed, thanks to the convexity of the objective function. As for the tuning parameter selection, we set $\lambda_{1m} = 2 \| \widehat{\boldsymbol{\Omega}}^{(0)}_m \|_{1, \infty} \sqrt{\frac{p_m \log p_m}{n p }} $ for mode-$m$, following \cite{li2022transfer}. For $\lambda_{2m}$, it is suggested to be determined via minimizing a BIC-type criterion, 
	$
	\frac{1}{2} \operatorname{tr} \{ \widehat{\boldsymbol{\Omega}}_{m}^{\top} \widehat{\boldsymbol{\Sigma}}_m^{\mathcal{A}} \widehat{\boldsymbol{\Omega}}_{m} \} -\operatorname{tr} \{ ( \widehat{\boldsymbol{\Delta}}_{m}^{\top}+ \boldsymbol{I}_{p_m} ) \widehat{\boldsymbol{\Omega}}_{m} \} + \frac{ \log N }{N} \| \widehat{\boldsymbol{\Omega}}_{m} \|_{0},
	$
	for each mode.

	\section{Statistical properties}\label{property}
	
	In this section, we establish some theoretical properties of the proposed transfer learning method. The following technical conditions are made.
	
	\begin{Con}\label{onenorm}
		For each $m \in [M]$ and $k \in [K]$, $\| \boldsymbol{\Omega}^*_m \|_{1, \infty}$ and $\| \boldsymbol{\Omega}^{(k)*}_m \|_{1, \infty}$ are bounded, and there is a constant $C_1$, satisfying $1/C_1 \leqslant \psi_{\min}(\boldsymbol{\Sigma}^*_m) \leqslant \psi_{\max}(\boldsymbol{\Sigma}^*_m)  \leqslant C_1$ and $1/C_1 \leqslant \psi_{\min}(\boldsymbol{\Sigma}^{(k)*}_m) \leqslant \psi_{\max}(\boldsymbol{\Sigma}^{(k)*}_m)  \leqslant C_1$.
	\end{Con}
	\begin{Con}\label{Irrep}
		Denote $\boldsymbol{\Gamma}_{m}^{*} = \boldsymbol{\Sigma}_{m}^{*} \otimes \boldsymbol{\Sigma}_{m}^{*}$, $S_{m} = \{ (i,j): [\boldsymbol{\Omega}^*_{m}]_{(i,j)} \ne 0 \}$, and $[\boldsymbol{\Gamma}_{m}^{*}]_{(S_m, S_m)}$ the sub-matrix with rows and columns of $\boldsymbol{\Gamma}_{m}^{*}$ indexed by $S_m$ and $S_m$, respectively. For each $m \in [M]$, $\| \boldsymbol{\Sigma}_{m}^{*} \|_{1, \infty}$ and $\| ([\boldsymbol{\Gamma}_{m}^{*}]_{(S_m, S_m)})^{-1} \|_{1, \infty}$ are bounded, and there exists some constant $C_2 \in (0,1]$ such that $\max_{e \in S_m^C} \| [\boldsymbol{\Gamma}_{m}^{*}]_{(e, S_m)}  ([\boldsymbol{\Gamma}_{m}^{*}]_{(S_m, S_m)})^{-1} \|_1 \leqslant 1 - C_2$.
	\end{Con}
	
	Condition \ref{onenorm} has been commonly assumed in the literature of Gaussian graphical models \citep{lam2009sparsistency,zhang2014sparse}. Condition \ref{Irrep} limits the influence of the non-connected terms in $S_m^C$ on the connected edges in $S_m$, which is also widely assumed to establish theoretical properties of lasso-type estimators \citep{ravikumar2011high,zhang2014sparse,lyu2019tensor}. Denote $\overline{p} = \max_{m \in [M]} p_m$, and $\overline{s} = \max_{m \in [M], j \in [p_m]} s_{mj}$ with $s_{mj} = \| \boldsymbol{\Omega}^*_{m(j)} \|_{0}$ that may diverge with $n$. We first state some existing result in \cite{lyu2019tensor}, which quantifies the asymptotic behavior of the initial estimate $\widehat{\boldsymbol{\Omega}}_m^{(0)}$.
	
	\begin{Lem}\label{cov}
		(\citealp{lyu2019tensor}) If Condition \ref{onenorm} holds, then $\| \widehat{\boldsymbol{\Sigma}}_m - \boldsymbol{\Sigma}_m^{*}\|_{\max} = O_p \left( \sqrt{\frac{p_m \log p_m}{n p }}  \right)$, for $m \in [M]$.
		If Condition \ref{Irrep} holds, $\overline{s} \sqrt{\frac{p_m \log p_m}{n p }} \ll 1$, and $p_1 \asymp \dots \asymp p_m$, then $\| \widehat{\boldsymbol{\Omega}}_m^{(0)} - \boldsymbol{\Omega}_m^{*}\|_{\max} = O_p \left( \sqrt{\frac{p_m \log p_m}{n p }}  \right)$.
		Furthermore, for $m \in [M]$, if the minimal signal of $\boldsymbol{\Omega}_m^{*}$ satisfies that $\sqrt{\frac{p_m \log p_m}{n p }} \lesssim \min_{(i,j) \in S_m } | [\boldsymbol{\Omega}_{m}^{*}]_{(i,j)}| $,
		then with probability tending to 1, $\widehat{S}_{m}^{(0)} = \{ (i,j): [\widehat{\boldsymbol{\Omega}}_{m}^{(0)}]_{(i,j)} \ne 0 \} = S_{m}$.
	\end{Lem}
	
	To describe the similarity between the target domain and auxiliary domains, we define the set of {\it informative} auxiliary domains as $\mathcal{A} = \{ k: \max_{m \in [M] } \{  \| \boldsymbol{\Delta}_{m}^{(k)*} \|_{1, \infty} + \| (\boldsymbol{\Delta}_{m}^{(k)*})^{\top} \|_{1, \infty} \} \leqslant h \}$ for a sufficiently small $h>0$. Clearly, $h$ measures the difference between precision matrices of each mode in the target and the $k$-th auxiliary domain.
	
	\subsection{All auxiliary domains are informative}
	
	We first consider an ideal scenario where all available auxiliary domains are informative.
	
	\begin{Con}\label{h-bound}
		Assume that $\mathcal{A} = [K]$.
	\end{Con}
	
	
	\begin{Mth}\label{step1}
		Suppose all the conditions of Lemma \ref{cov} and Condition \ref{h-bound} are met, 
		$n \leqslant N$ with $N = \sum_{k=1}^{K} n_k$, and $\lambda_{1m} = C (1+h) \sqrt{\frac{\overline{p} \log \overline{p}}{n p }}$ for a sufficiently large constant $C$. For $\widehat{\boldsymbol{\Sigma}}_m^{\mathcal{A}}$ in (\ref{na-weight}), it holds true that $\| \widehat{\boldsymbol{\Delta}}_m - \boldsymbol{\Delta}^*_m \|_{2, \infty}^2 = O_p ( \delta_{h})$ for $m \in [M]$, where $\delta_{h} = (1+h) h \sqrt{\frac{\overline{p} \log \overline{p}}{n p }} \wedge h^2$.
	\end{Mth}
	\begin{Mth}\label{step2}
		If the conditions of Theorem \ref{step1} hold, and $\lambda_{2m} = C \left( \sqrt{\frac{\delta_{h}}{\overline{s}}} + \sqrt{\frac{\overline{p} \log \overline{p}}{N p }} \right) $ for a sufficiently large constant $C$, then $\| \widehat{\boldsymbol{\Omega}}_{m} -  \boldsymbol{\Omega}_{m}^* \|_{2, \infty}^2 \vee \frac{1}{p_m} \| \widehat{\boldsymbol{\Omega}}_{m} -  \boldsymbol{\Omega}_{m}^* \|_{F}^2= O_p \left( \frac{\overline{s} \overline{p} \log \overline{p}}{(N + n)p } + \delta_{h} \right)$ for $m \in [M]$.
	\end{Mth}
	\begin{Rem}\label{theta0}
		Note that Lemma \ref{cov} implies that $\| \widehat{\boldsymbol{\Omega}}^{(0)}_m - \boldsymbol{\Omega}^*_m \|_{2, \infty}^2 \vee \frac{1}{p_m} \| \widehat{\boldsymbol{\Omega}}^{(0)}_m - \boldsymbol{\Omega}^*_m \|_{F}^2 = O_p ( \frac{\overline{s} \overline{p} \log \overline{p}}{ n p} )$ for $m \in [M]$. It is thus clear that the proposed transfer learning method achieves a faster convergence rate when $N \gg n$ and $h \ll \overline{s} \sqrt{\frac{\overline{p} \log \overline{p}}{n p }}$. 
	\end{Rem}
	
	We next establish the variable selection consistency of the proposed transfer learning method in terms of exactly recovering the tensor graphical model. Some additional conditions are necessary.
	\begin{Con}\label{Irrep-a}
		Define $\boldsymbol{\Sigma}_m^{\mathcal{A}*} = \sum_{k=1}^{K} \alpha_k \boldsymbol{\Sigma}_m^{(k)*}$ with $\sum_{k=1}^{K} \alpha_k = 1$, then for each $m \in [M]$, $\| \boldsymbol{\Sigma}_{m}^{\mathcal{A}*} \|_{1, \infty}$ and $\max_{j \in [p_m]} \| ([ \boldsymbol{\Sigma}_{m}^{\mathcal{A}*} ]_{(S_{mj},S_{mj})})^{-1} \|_{1, \infty}$ are bounded, and there exists some constant $C_{3} \in (0,1]$ such that $\max_{j \in [p_m], e \in S_{mj}^C} \| [\boldsymbol{\Sigma}_m^{\mathcal{A}*}]_{(e, S_{mj})}  ([\boldsymbol{\Sigma}_m^{\mathcal{A}*}]_{(S_{mj}, S_{mj})})^{-1} \|_1 \leqslant 1 - C_{3}$, where $S_{mj} = \{ i \in [p_m]: [\boldsymbol{\Omega}^*_{m}]_{(i,j)} \ne 0 \}$ and $S_{mj}^C = \{ i \in [p_m]: [\boldsymbol{\Omega}^*_{m}]_{(i,j)} = 0 \}$. 
	\end{Con}
	\begin{Con}\label{h0-bound}
		Assume that $\max_{m \in [M], k \in [K] } \| \boldsymbol{\Delta}_{m}^{(k)*} \|_{\max} \lesssim h/\overline{s}$.
	\end{Con}
	Condition 4 imposes the irrepresentability condition on the auxiliary domains, to quantify the behavior of $\mathcal{Q}_{2} ( \boldsymbol{\Omega}_{m} )$ in Step 2(b).
	Condition \ref{h0-bound} is necessary for establishing estimation error of $\widehat{\boldsymbol{\Omega}}_{m}$ in max norm, which is mild due to the fact that $\overline{s} < \overline{p}$.
	\begin{Mth}\label{step2-sp}
		If the conditions of Theorem \ref{step2} and Conditions \ref{Irrep-a} to \ref{h0-bound} hold, and $h$ is bounded, then $\| \widehat{\boldsymbol{\Omega}}_{m} -  \boldsymbol{\Omega}_{m}^* \|_{\max}  = O_p \left( \sqrt{\frac{\delta_{h}}{\overline{s}}} + \sqrt{\frac{\overline{p} \log \overline{p}}{(N+n) p }} \right)$ for $m \in [M]$. Furthermore, if $ \sqrt{\frac{\delta_{h}}{\overline{s}}} + \sqrt{\frac{\overline{p} \log \overline{p}}{(N+n) p }}  \lesssim  \min_{(i,j) \in S_m } | [\boldsymbol{\Omega}_{m}^{*}]_{(i,j)}|$, 
		then with probability tending to 1, $\widehat{S}_{m} = \{ (i,j): [\widehat{\boldsymbol{\Omega}}_{m}]_{(i,j)} \ne 0 \} = S_{m}$ for $m \in [M]$.
	\end{Mth}
	It is interesting to note that $\sqrt{\frac{\delta_{h}}{\overline{s}}} + \sqrt{\frac{\overline{p} \log \overline{p}}{(N+n) p }} \ll \sqrt{\frac{\overline{p} \log \overline{p}}{n p }} $, if $N \gg n$ and $h \ll \overline{s} \sqrt{\frac{\overline{p} \log \overline{p}}{n p }}$, thus it can be concluded that the minimum signal condition required for eatablishing the variable selection consistency for the proposed transfer learning method is much weaker than that when using the target domain only.

	\subsection{At least one informative auxiliary domain}
	
	We now turn to a more complex case where some non-informative auxiliary domains dominates, so that the model selection step may force the final estimator to become the initial estimate based on the target domain only, and then another part of information on the informative auxiliary domains will be offset. At this point, it only ensures that the transfer learning does not deteriorate, but does not make full use of positive information from informative auxiliary domains. Therefore, we further consider the theoretical properties of the proposed method based on $\widehat{\boldsymbol{\Sigma}}_m^{\mathcal{A}}$ constructed by the data-adaptive weights.
	
	\begin{Con}\label{h-bound-o}
		There exists a $h \lesssim \overline{s} \sqrt{\frac{\overline{p} \log \overline{p}}{n p }}$ such that the positive set $\mathcal{A} \subseteq [K]$ is non-empty.
	\end{Con}
	\begin{Con}\label{Irrep-a-o}
		Re-define $\boldsymbol{\Sigma}_m^{\mathcal{A}*} = \sum_{k \in \mathcal{A}} \alpha_k \boldsymbol{\Sigma}_m^{(k)*}$ with $\sum_{k\in \mathcal{A}} \alpha_k = 1$, then for each $m \in [M]$, $\| \boldsymbol{\Sigma}_{m}^{\mathcal{A}*} \|_{1, \infty}$ and $\max_{j \in [p_m]} \| ([ \boldsymbol{\Sigma}_{m}^{\mathcal{A}*} ]_{(S_{mj},S_{mj})})^{-1} \|_{1, \infty}$ are bounded, and there exists some constant $C_{3} \in (0,1]$ such that $\max_{j \in [p_m], e \in S_{mj}^C} \| [\boldsymbol{\Sigma}_m^{\mathcal{A}*}]_{(e, S_{mj})}  ([\boldsymbol{\Sigma}_m^{\mathcal{A}*}]_{(S_{mj}, S_{mj})})^{-1} \|_1 \leqslant 1 - C_{3}$, where $S_{mj} = \{ i \in [p_m]: [\boldsymbol{\Omega}^*_{m}]_{(i,j)} \ne 0 \}$ and $S_{mj}^C = \{ i \in [p_m]: [\boldsymbol{\Omega}^*_{m}]_{(i,j)} = 0 \}$. 
	\end{Con}
	\begin{Con}\label{h0-bound-o}
		Assume that $\max_{m \in [M], k \in \mathcal{A} } \| \boldsymbol{\Delta}_{m}^{(k)*} \|_{\max} \lesssim h/\overline{s}$.
	\end{Con}
	
	Conditions \ref{h-bound-o} to \ref{h0-bound-o} are weakened  forms of Conditions \ref{h-bound} to \ref{h0-bound}, respectively, in which the assumption of similarity is only imposed on informative auxiliary domains.
	
	\begin{Mth}\label{step2-o}
		Suppose all the conditions of Lemma \ref{cov} and Condition \ref{h-bound-o} are met, $n_1 \asymp \dots \asymp n_K$, $n \leqslant N_{\mathcal{A}}$ with $N_{\mathcal{A}} = \sum_{k \in \mathcal{A}} n_k$, $K = O(1)$, $\lambda_{1m} = C (1+h) \sqrt{\frac{\overline{p} \log \overline{p}}{n p }}$, and $\lambda_{2m} = C \left( \sqrt{\frac{\delta_{h}}{\overline{s}}} + \sqrt{\frac{\overline{p} \log \overline{p}}{ p N_{\mathcal{A}} }} \right) $ for a sufficiently large constant $C$.
		For $\widehat{\boldsymbol{\Sigma}}_m^{\mathcal{A}}$ in (\ref{adap-weight}), it holds true that
		$\| \widehat{\boldsymbol{\Omega}}_{m} -  \boldsymbol{\Omega}_{m}^* \|_{2, \infty}^2 \vee \frac{1}{p_m} \| \widehat{\boldsymbol{\Omega}}_{m} -  \boldsymbol{\Omega}_{m}^* \|_{F}^2= O_p \left( \frac{\overline{s} \overline{p} \log \overline{p}}{(N_{\mathcal{A}} + n)p } + \delta_{h} \right)$ for $m \in [M]$, where $\delta_{h} = (1+h) h \sqrt{\frac{\overline{p} \log \overline{p}}{n p }} \wedge h^2$. 
		Furthermore, if Conditions \ref{Irrep-a-o} to \ref{h0-bound-o} hold, and assume that $ \sqrt{\frac{\delta_{h}}{\overline{s}}} + \sqrt{\frac{\overline{p} \log \overline{p}}{(N_{\mathcal{A}}+n) p }}  \lesssim  \min_{(i,j) \in S_m } | [\boldsymbol{\Omega}_{m}^{*}]_{(i,j)}|$, then with probability tending to 1, $\| \widehat{\boldsymbol{\Omega}}_{m} -  \boldsymbol{\Omega}_{m}^* \|_{\max}  = O_p \left( \sqrt{\frac{\delta_{h}}{\overline{s}}} + \sqrt{\frac{\overline{p} \log \overline{p}}{(N_{\mathcal{A}}+n) p }} \right)$ and $\widehat{S}_{m} = S_{m}$ for $m \in [M]$.
	\end{Mth}
	
	It is clear that as long as there is at least one informative auxiliary domain, satisfying $N_{\mathcal{A}} \gg n$ and $h \ll \overline{s} \sqrt{\frac{\overline{p} \log \overline{p}}{n p }}$, the proposed transfer learning method based on data-adaptively defined $\widehat{\boldsymbol{\Sigma}}_m^{\mathcal{A}}$ can improve estimation errors benefiting from its information, and is not affected by the possible presence of non-informative auxiliary domains, which shows the powerful robustness to complex scenarios.
	
	\begin{Rem}\label{step3}
		If the ideal assumption about informative auxiliary domains is violated in practice, the transfer learning may be counterproductive. As suggested in \cite{li2022transfer}, the selection step (\ref{omega_obj}) can theoretically guarantee that the final estimator $\widehat{\boldsymbol{\Omega}}_{m}^{(f)}$ is as effective as $\widehat{\boldsymbol{\Omega}}_m$ if $h \lesssim \overline{s} \sqrt{\frac{\overline{p} \log \overline{p}}{n p }}$, and $\widehat{\boldsymbol{\Omega}}_{m}^{(f)}$ is still no less effective than $\widehat{\boldsymbol{\Omega}}^{(0)}_m$ if $h \gg \overline{s} \sqrt{\frac{\overline{p} \log \overline{p}}{n p }}$ for $\widehat{\boldsymbol{\Sigma}}_m^{\mathcal{A}}$ in (\ref{na-weight}), or the informative set $\mathcal{A}$ is empty for $\widehat{\boldsymbol{\Sigma}}_m^{\mathcal{A}}$ in (\ref{adap-weight}).
	\end{Rem}

	\section{Simulation}
	
	We consider two types of target graphs.
	\begin{itemize}
		\item Chain graph. For each $m \in [M]$, the $(i,j)$-th entry of $\boldsymbol{\Omega}_m^*$ is set as 1 if $i=j$; $\exp (- \rho_{ij} /2)$ with $\rho_{ij} = \rho_{ji}$ generated from $\mathrm{Unif}(0.5,1)$, if $|i - j|=1$; and 0, if $|i - j|>1$.
		\item Nearest neighbor graph. For each $m \in [M]$, we randomly generate $p_m$ points in a unit square and locate four nearest neighbors for each point. The corresponding entries in $\boldsymbol{\Omega}_m$ are uniformly sampled from $[-1,-0.5] \cup [0.5,1]$. The final precision matrix is generated as $\boldsymbol{\Omega}_m^* = \boldsymbol{\Omega}_m + |\psi_{\min}(\boldsymbol{\Omega}_m) + 0.2| \boldsymbol{I}_{p_m}$ to ensure the positive definiteness.
	\end{itemize}
	For each target graph, we set $M=3$ with dimensions $(p_1, p_2, p_3) = (10, 10, 20)$ or $M=2$ with dimensions $(p_1, p_2) = (100, 100)$, and set the size of the target graph as $n=50$. We also consider two different simulation scenarios.
	
	\textbf{Scenario 1.} We consider $\mathcal{A} = [K]$ and vary $K \in \{ 1, \cdots, 5 \}$, that is, all auxiliary domains are informative with size $n_k = 80$ for $k \in [K]$, where $[ \boldsymbol{\Delta}_m^{(k)} ]_{(i,j)}=0$ with probability 0.9 or randomly generated from $\mathrm{Unif} [- h_{01}, h_{01}]$ with probability 0.1, and $h_{01} = \sqrt{\frac{\overline{p} \log \overline{p}}{n p }}$.
	
	\textbf{Scenario 2.} We fix $K=5$ with size $n_k = 100$ for $k \in [K]$ and vary $\mathrm{card}(\mathcal{A}) \in \{ 0, 1, \cdots, K \}$. The informative auxiliary domains with $k \in \mathcal{A}$ are generated similarly as Scenario 1. For $k \notin \mathcal{A}$, $[ \boldsymbol{\Delta}_m^{(k)} ]_{(i,j)}=0$ with probability 0.75, or randomly generated from $\mathrm{Unif} [- h_{02}, h_{02}]$ with probability 0.25, where $h_{02} = 10 \overline{s} \sqrt{\frac{\overline{p} \log \overline{p}}{n p }}$.
	
	We compare three methods in Scenario 1, including the single task tensor graphical model using the target domain only, which is implemented in the R package ``Tlasso", and the proposed methods with the auxiliary covariance in (\ref{na-weight}) and (\ref{adap-weight}),  denoted as ``proposed" and ``proposed.v", respectively. In Scenario 2, we further consider another ``oracle" method, which applies ``proposed" on the target domain and the known informative auxiliary domains.
	
	The performances of the competing methods are measured by a number of metrics: (1) estimation error in Frobenius norm of Kronecker product of precision matrices, defined as $\| \widehat{\boldsymbol{\Omega}}^{(K)} - \boldsymbol{\Omega}^{(K)*} \|_F$, where 
	$\widehat{\boldsymbol{\Omega}}^{(K)} = \widehat{\boldsymbol{\Omega}}_1^{(f)} \otimes \cdots \otimes \widehat{\boldsymbol{\Omega}}_M^{(f)}$ and $\boldsymbol{\Omega}^{(K)*} = \boldsymbol{\Omega}_1^* \otimes \cdots \otimes \boldsymbol{\Omega}_M^*$; (2) averaged estimation errors in Frobenius norm of all modes $\frac{1}{M} \sum_{m=1}^{M} \| \widehat{\boldsymbol{\Omega}}^{(f)}_m - \boldsymbol{\Omega}^{*}_m \|_F$; (3) averaged estimation errors in max norm of all modes $\frac{1}{M} \sum_{m=1}^{M} \| \widehat{\boldsymbol{\Omega}}^{(f)}_m - \boldsymbol{\Omega}^{*}_m \|_{\max}$; (4) the true positive rate (TPR) and the true negative rate (TNR) of the Kronecker product of precision matrices; (5) the averaged TPRs and TNRs of all modes. All metrics are averaged based on 100 independent replications.
	
\begin{figure}[!htb]
	\centering
	\subfigure{\includegraphics[scale = 0.24]{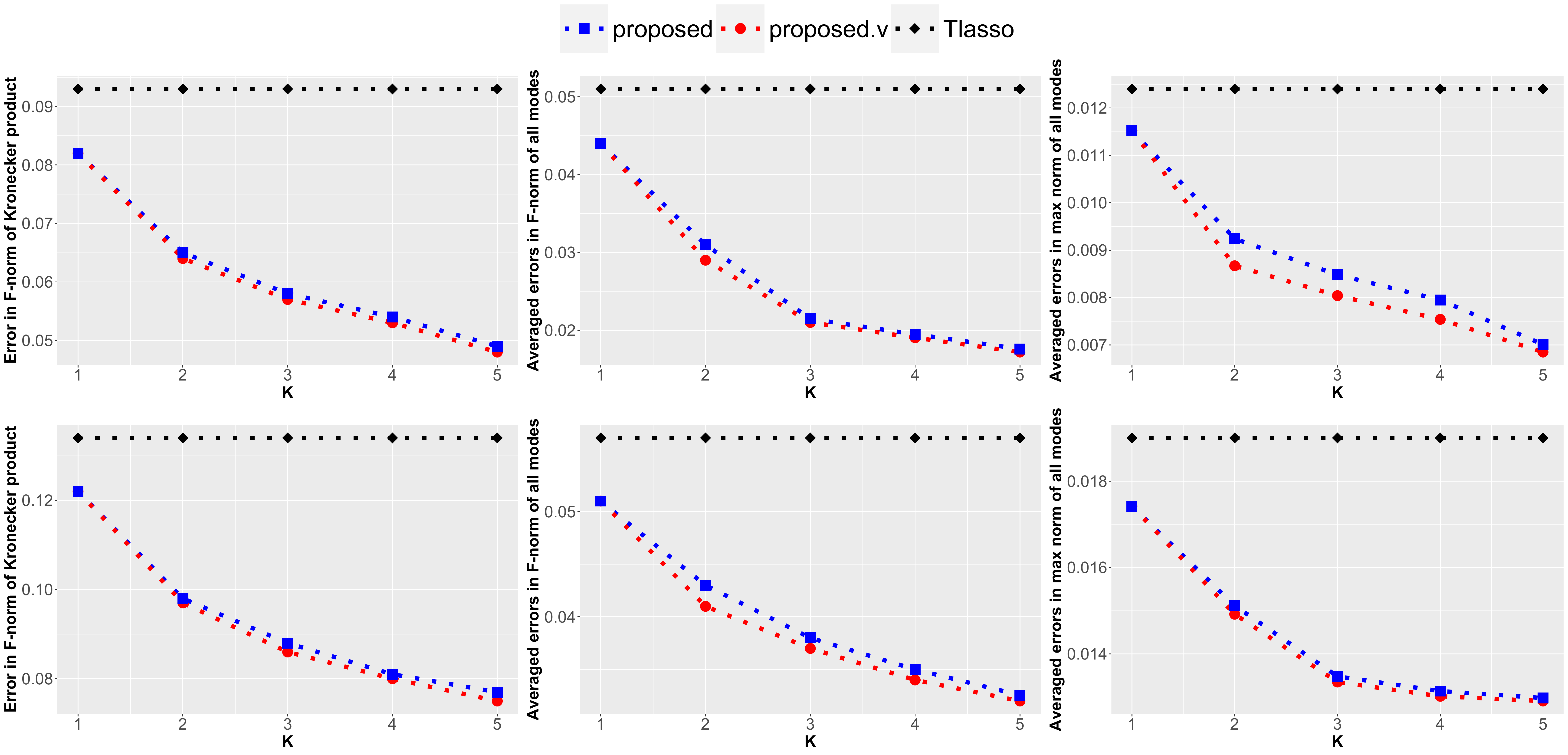}}
	\caption{Averaged metrics of estimation errors over 100 replications for Scenario 1 with $M=3$. The top and bottom rows correspond to the chain and nearest neighbor graph, respectively.}
	\label{a_M3}
\end{figure}

\begin{figure}[!htb]
	\centering
	\subfigure{\includegraphics[scale = 0.24]{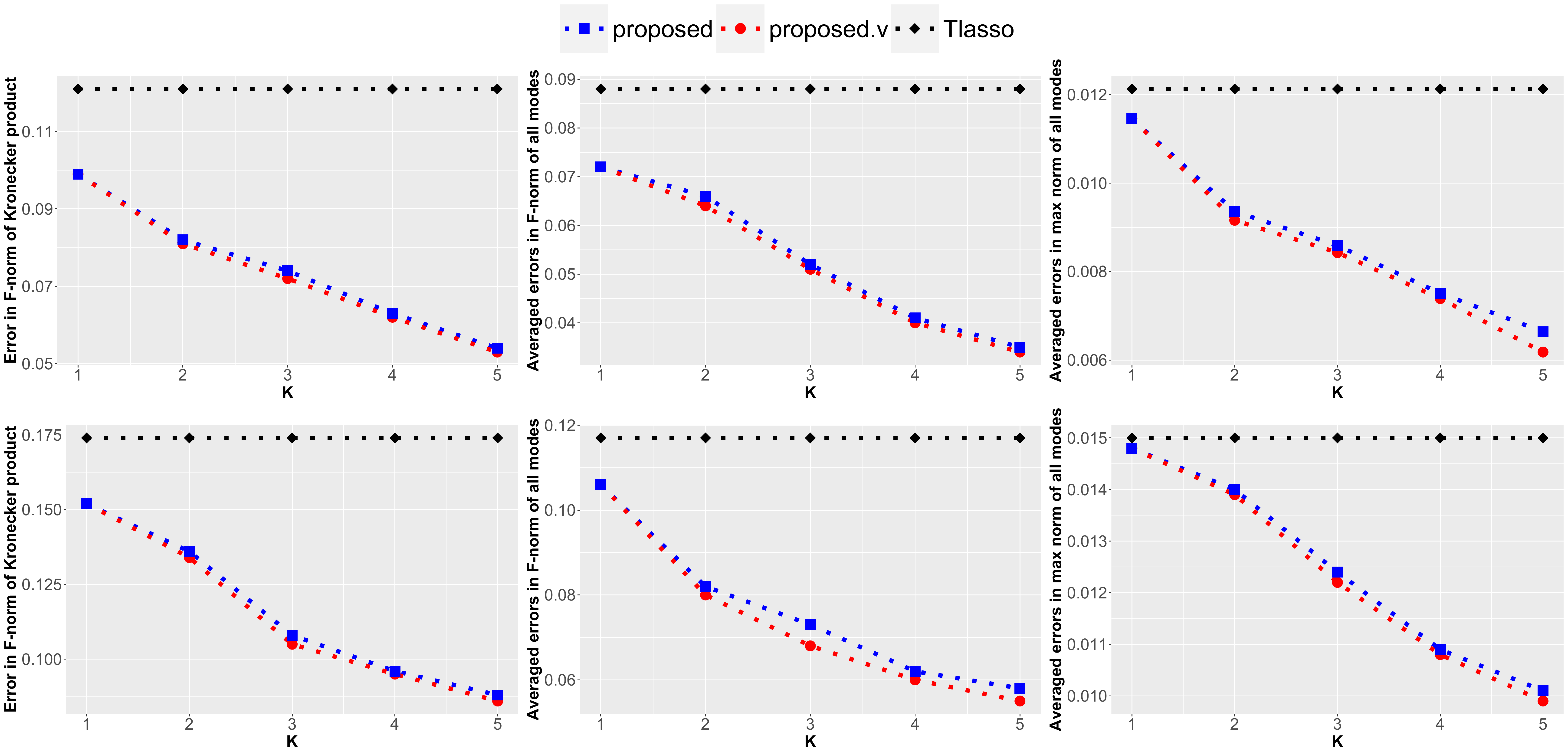}}
	\caption{Averaged metrics of estimation errors over 100 replications for Scenario 1 with $M=2$. The top and bottom rows correspond to the chain and nearest neighbor graph, respectively.}
	\label{a_M2}
\end{figure}

\begin{figure}[!htb]
	\centering
	\subfigure{\includegraphics[scale = 0.25]{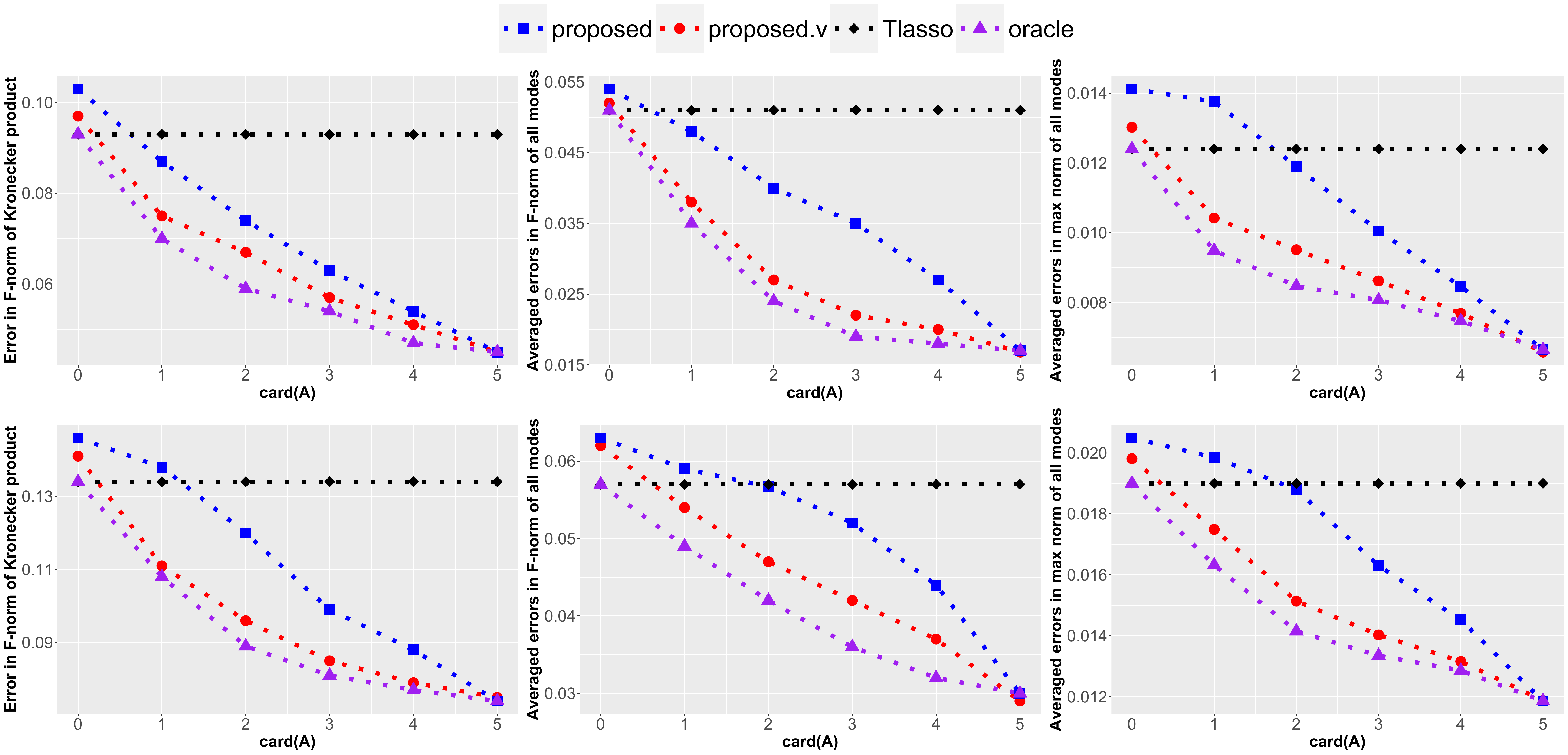}}
	\caption{Averaged metrics of estimation errors over 100 replications for Scenario 2 with $M=3$. The top and bottom rows correspond to the chain and nearest neighbor graph, respectively.}
	\label{M3}
\end{figure}

\begin{figure}[!htb]
	\centering
	\subfigure{\includegraphics[scale = 0.25]{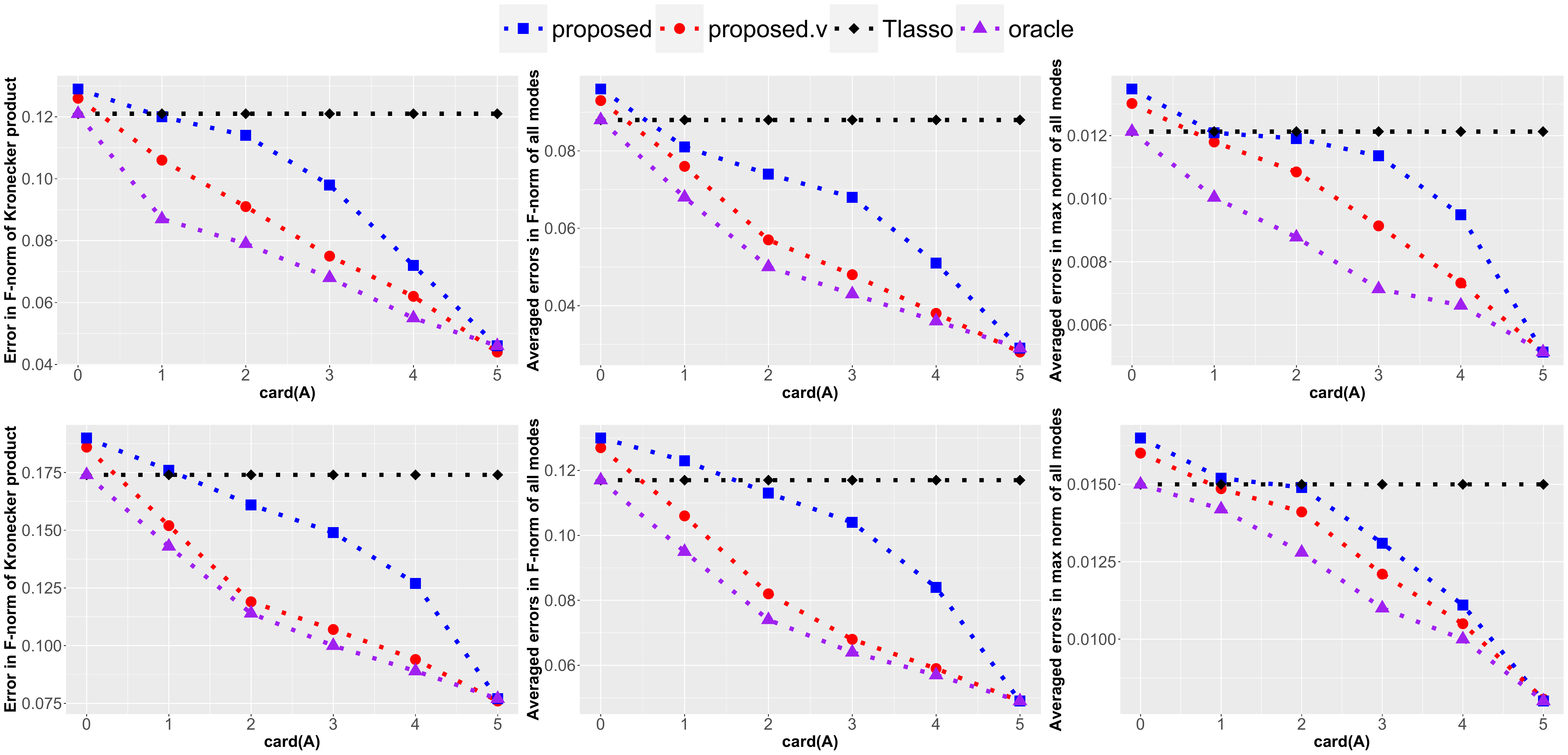}}
	\caption{Averaged metrics of estimation errors over 100 replications for Scenario 2 with $M=2$. The top and bottom rows correspond to the chain and nearest neighbor graph, respectively.}
	\label{M2}
\end{figure}

	The first three estimation errors are summarized in Figures \ref{a_M3} to \ref{M2}, whereas the parameter selection metrics are summarized in Supporting Information. Observations made under different settings are very similar. For example, in Scenario 1 where all auxiliary domains are informative, as the number of auxiliary domains $K$ increases, all estimation errors of the two proposed transfer learning-based methods decrease with no significant difference from each other and both are better than Tlasso as expected. In Scenario 2, the two proposed methods are not significantly inferior to Tlasso thanks to the model selection step, when there is no informative auxiliary domain. It is interesting to remark that the two proposed methods have different performance paths as the number of informative auxiliary domains $\mathrm{card}(\mathcal{A})$ increases. Specifically, the estimation errors of ``proposed.v", whose weights are constructed based on both sample sizes and differences between the target and auxiliary domains, decrease so fast that it can dominate Tlasso even when there is only one informative auxiliary domain, and its overall performance is comparable to ``oracle". However, ``proposed" is more affected by the non-informative auxiliary domains, whose estimation errors are much larger than ``proposed.v" and sometimes only outperform Tlasso when there are relatively large number of informative auxiliary domains. As for the performances of variable selection, all methods have achieved 100\% TPR in all settings, and the TNRs of the two proposed methods are significantly improved compared with Tlasso, thanks to the informative auxiliary domains.

	\section{ADHD brain functional networks}
	
	In this section, we apply the proposed method to study functional connectivity behaviors among brain regions in the attention deficit hyperactivity disorder (ADHD) disease datasets across multiple sites. In the brain functional network, typically, a node corresponds to an anatomically defined brain region, and the present of connectivity between a pair of nodes to a measure of inter-regional dependency. Resting-state functional magnetic resonance imaging (rs-fMRI) is widely used to measure spontaneous low-frequency blood oxygen level dependent (BOLD) signal fluctuations within several minutes in some brain regions, so that the functional synchronization of brain systems, that is, the connections of the brain network, can be characterized. There is growing evidence that the brain functional connectivity network is altered in response to ADHD and is important to explore the pathogenesis and diagnosis, while Gaussian graphical model is an important statistical tool to detect this brain functional connectivity \citep{zhu2018multiple}.

	The analyzed dataset is publicly available at the ADHD-200 preprocessed repository \citep{bellec2017neuro}. This dataset is collected from seven sites, each of which contains demographical information, phenotypic data, and rs-fMRI of two groups consisting of typically developing controls (TDC) and ADHD, and is pre-processed following the standard Athena pipeline \citep{bellec2017neuro}. Only those rs-fMRI scans that pass the quality control are included in our analysis. The names of the seven sites and their sample sizes of TDC and ADHD groups are summarized in Table \ref{test}.
	The brain image is parcellated into 116 regions of interet (ROIs) following the Anatomical Automatic Labeling (AAL) atlas, and each ROI is recorded with the BOLD signal fluctuations at $T$ time points. Therefore, the rs-fMRI from each subject takes the form of a $T \times 116$-dimensional tensor. To explore brain functional connectivity, we are only interested in the second mode, that is, the spatial mode corresponding to ROIs. Here we note that although the first mode, which is the temporal mode corresponding to the time series, is not the target of the analysis, its existence leads to the necessity of tensor instead of multivariate Gaussian graphical model \citep{zhu2018multiple}.

	To compare competitors fairly and comprehensively, we rotated one site as the target domain and other sites as auxiliary domains, and TDC and ADHD groups are considered separately. Note that the underlying true parameters of precision matrices are unavailable, so we use the negative log-likelihood based on five-fold cross-validation as an indicator to evaluate the performance of all competitors when a site is fixed as the target domain. Specifically, samples of the target domain are randomly divided into five parts, one of which is used as the test sample to calculate the covariance matrices of all modes $\{ \widehat{\boldsymbol{\Sigma}}_m^{\mathrm{test}} \}_{m=1}^{M}$ and the rest is the training sample. The out-of-sample absolute prediction error of an arbitrary estimator $\widehat{\boldsymbol{\Omega}}_m^{o}$ for the $m$-th mode, estimated using the training sample, is defined as 
	\begin{equation}\nonumber
		\begin{aligned}
			\mathrm{PE} (\widehat{\boldsymbol{\Omega}}_m^{o} ) = -\frac{1}{p_m} \log [\det (\widehat{\boldsymbol{\Omega}}_m^{o})] + \frac{1}{p_m} \operatorname{tr} (\widehat{\boldsymbol{\Sigma}}_m^{\mathrm{test}} \widehat{\boldsymbol{\Omega}}_m^{o} ).
		\end{aligned}
	\end{equation}
	Note that the negative log-likelihood are widely used to evaluate method effectiveness for unsupervised graph model problems \citep{li2022transfer}, especially when the underlying network structure is unknown. For the estimator of proposed method $\widehat{\boldsymbol{\Omega}}_m$ and its variant $\widehat{\boldsymbol{\Omega}}_m^{v}$, their relative prediction errors are defined as $\frac{ \mathrm{PE}(\widehat{\boldsymbol{\Omega}}_m ) }{\mathrm{PE}(\widehat{\boldsymbol{\Omega}}_m^{(0)} )}$ and $\frac{ \mathrm{PE}(\widehat{\boldsymbol{\Omega}}_m^{v} ) }{\mathrm{PE}(\widehat{\boldsymbol{\Omega}}_m^{(0)} )}$, respectively, where $\widehat{\boldsymbol{\Omega}}_m^{(0)}$ is Tlasso estimator. Here we only consider the $m=2$-th mode corresponding to ROIs of interest. Average errors of five-fold cross-validation are summarized in Table \ref{test}. It is clear that the two proposed methods outperform Tlasso under almost all sites as target domains, and the relative performances of the proposed methods are improved by more than 10\% in about half the cases. 

\begin{figure}[!htb]
	\centering
	\subfigure{\includegraphics[scale = 0.31]{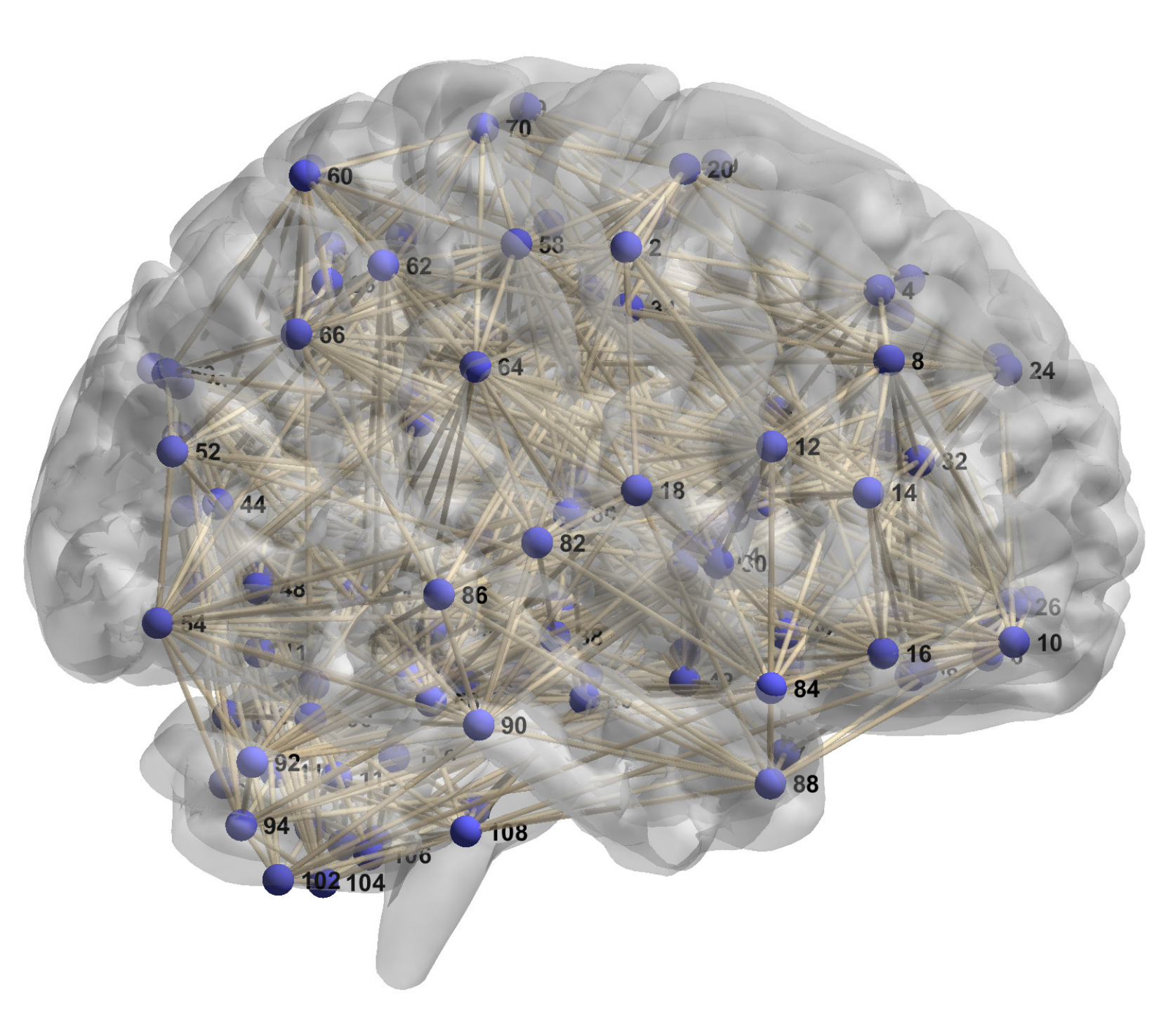}} \ 
	\subfigure{\includegraphics[scale = 0.29]{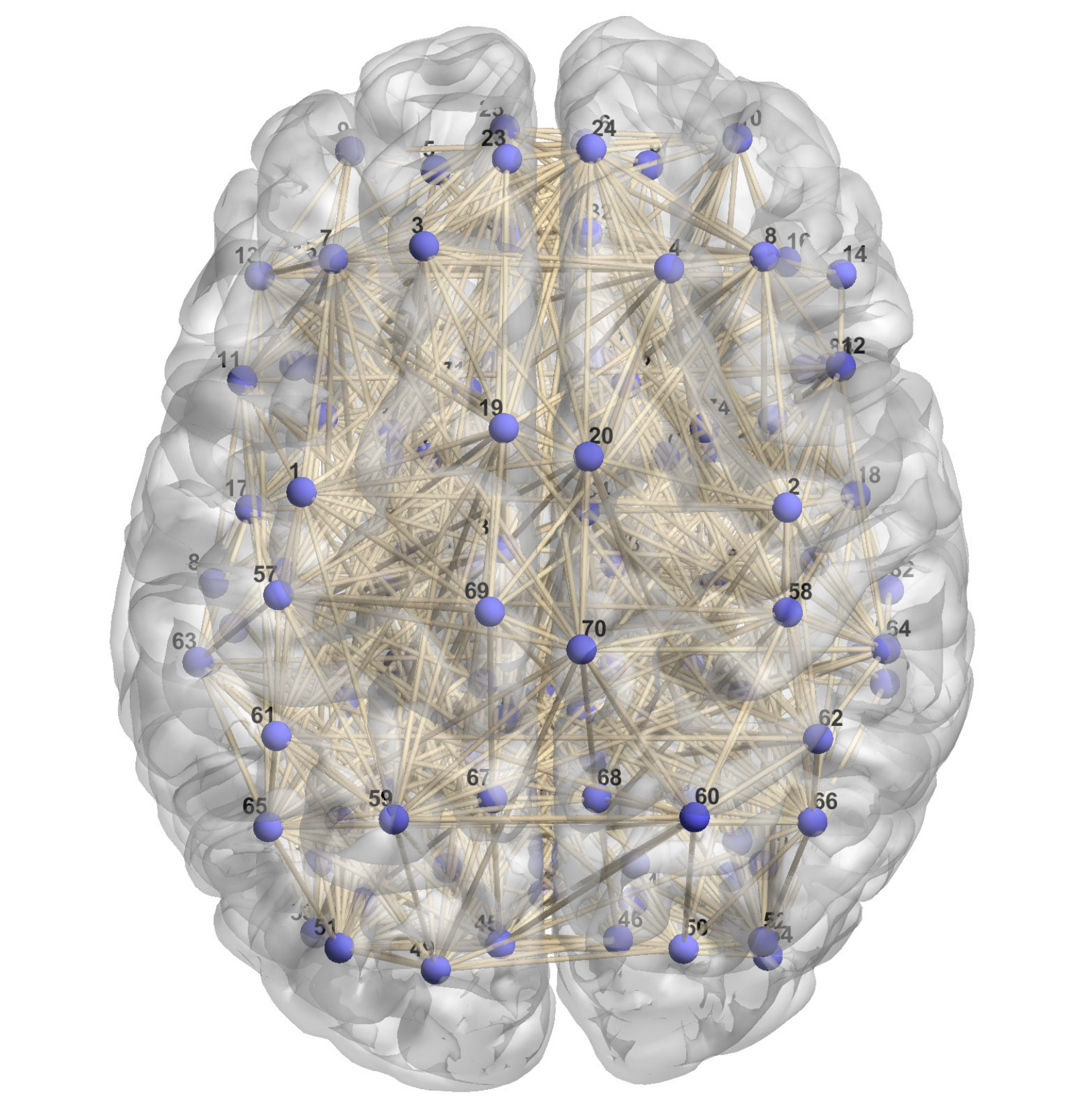}} \
	\subfigure{\includegraphics[scale = 0.29]{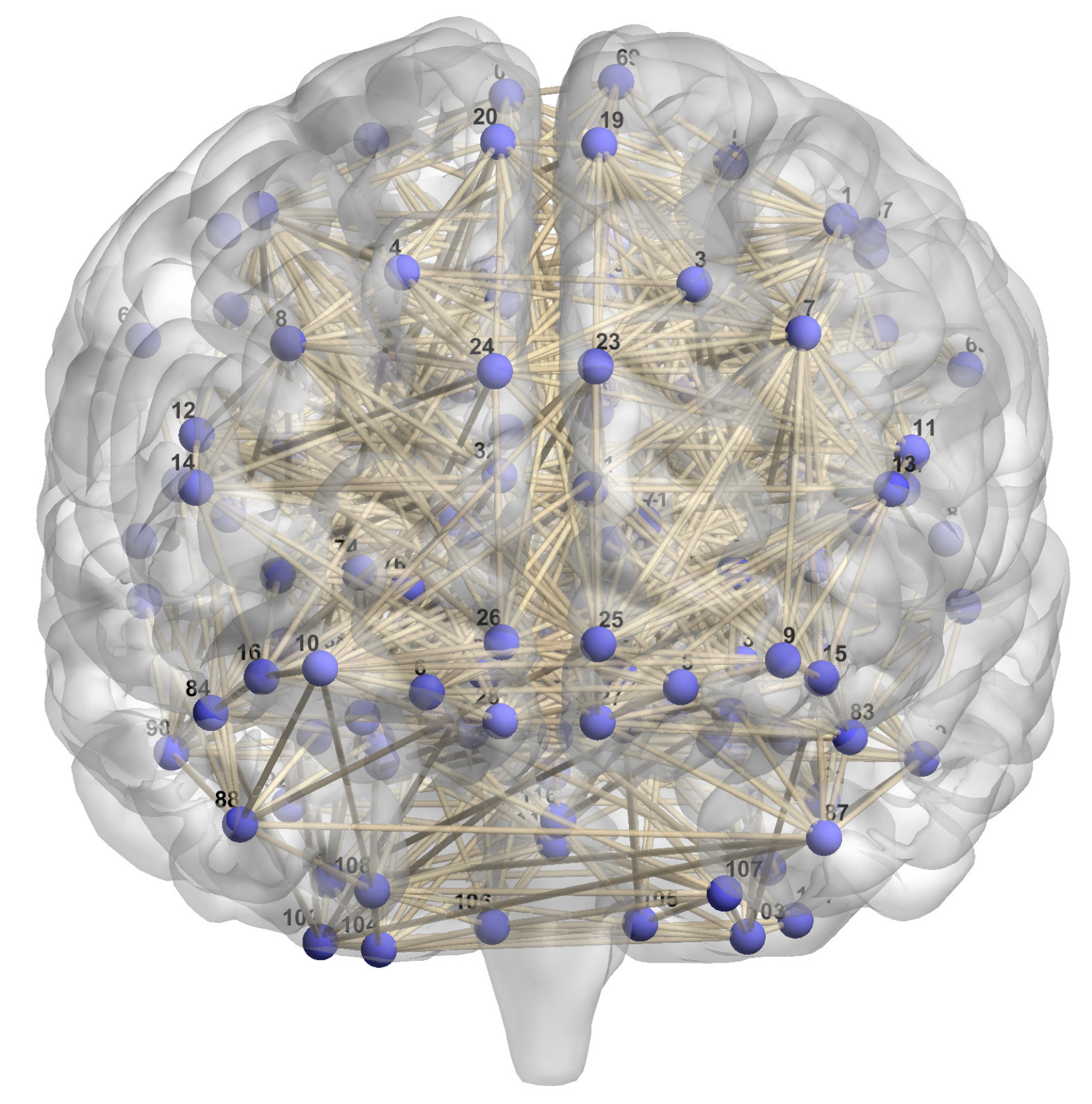}}
	\subfigure{\includegraphics[scale = 0.31]{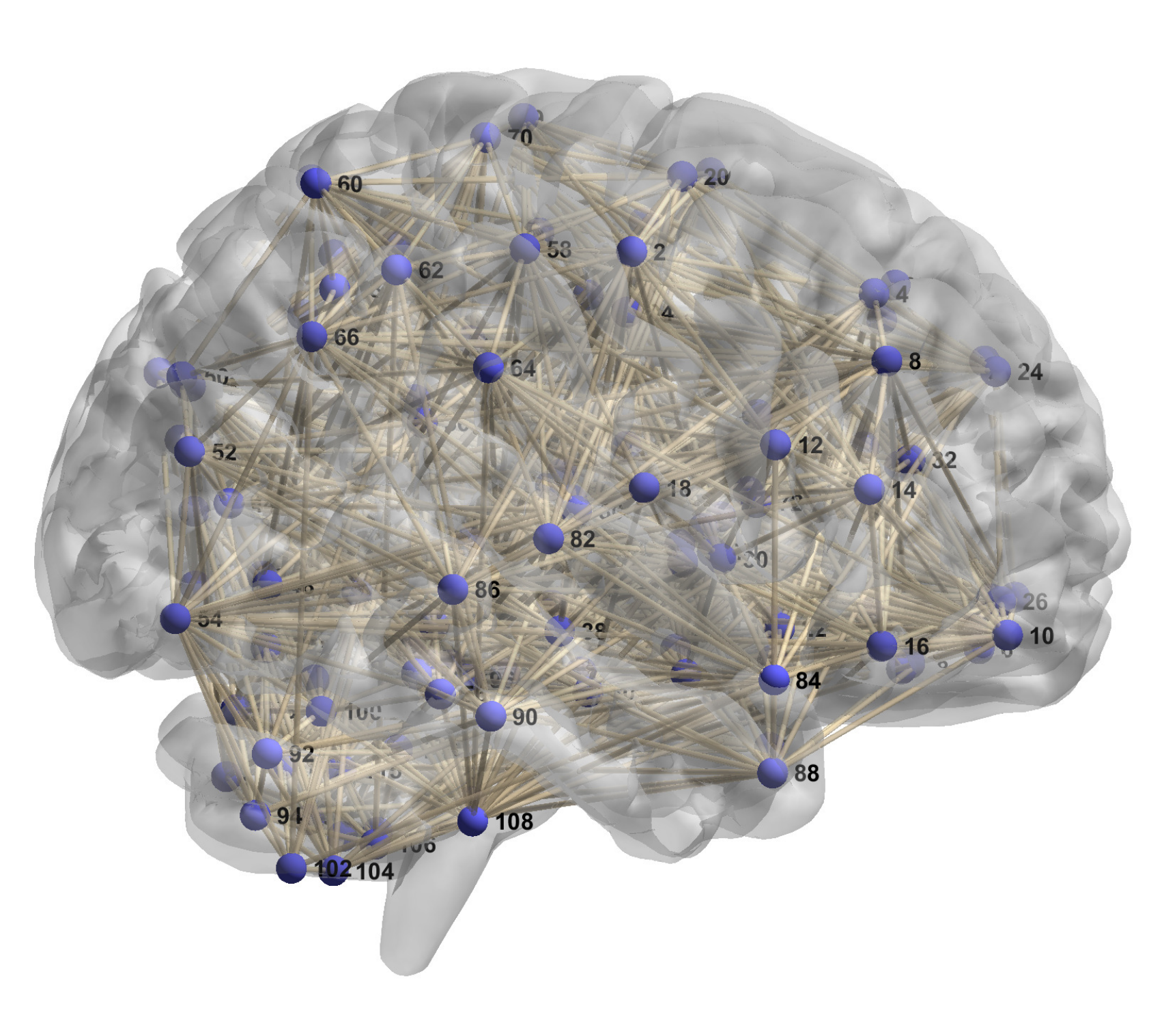}}
	\subfigure{\includegraphics[scale = 0.29]{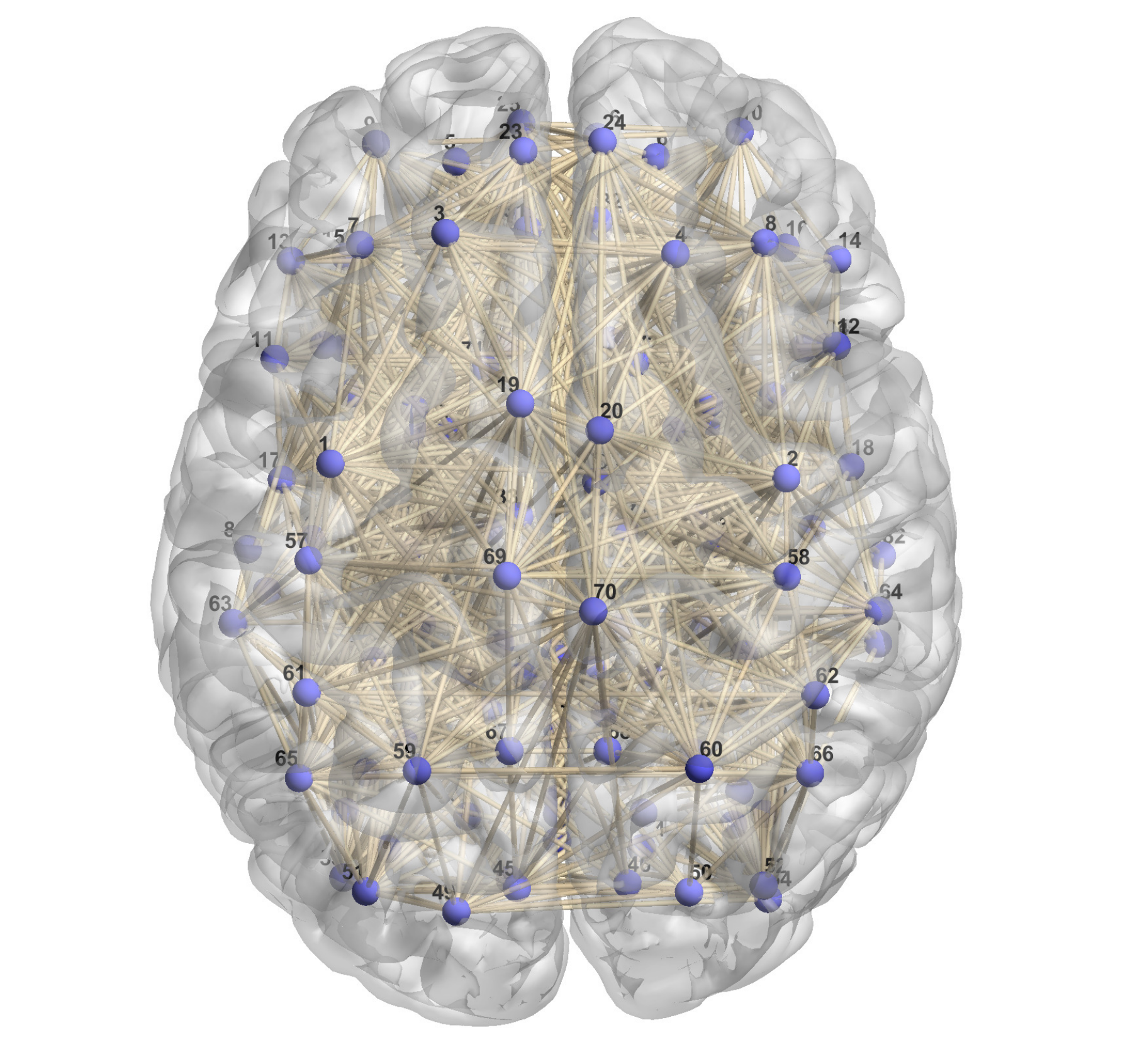}}
	\subfigure{\includegraphics[scale = 0.29]{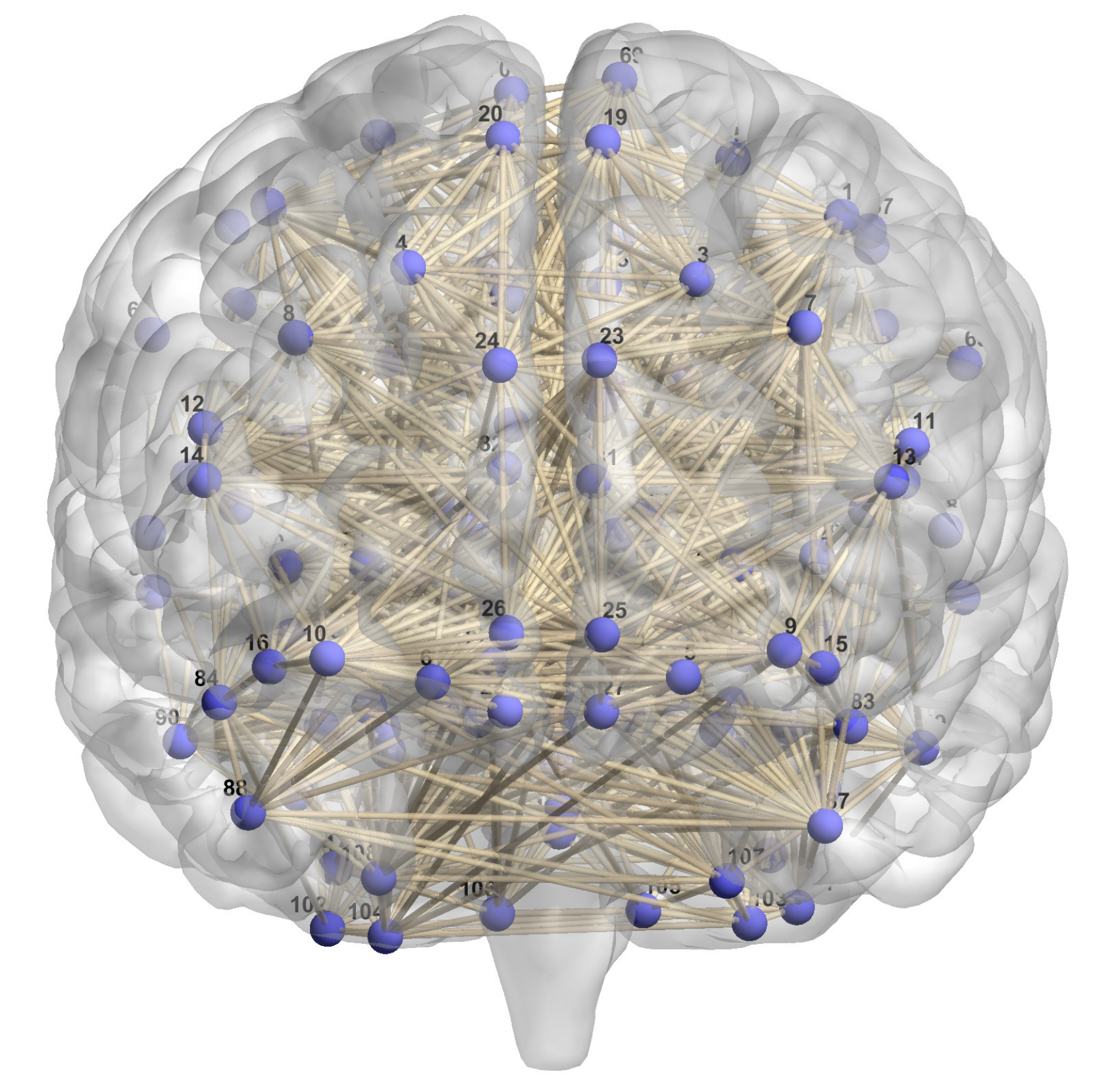}}
	\caption{The networks of brain functional connectivity of TDC (top) and ADHD (bottom) groups. In each row, three different views are also provided: sagittal (left), axial (middle), and coronal (right).}
	\label{abrain}
\end{figure}

\begin{figure}[!htb]
	\centering
	\subfigure{\includegraphics[scale = 0.31]{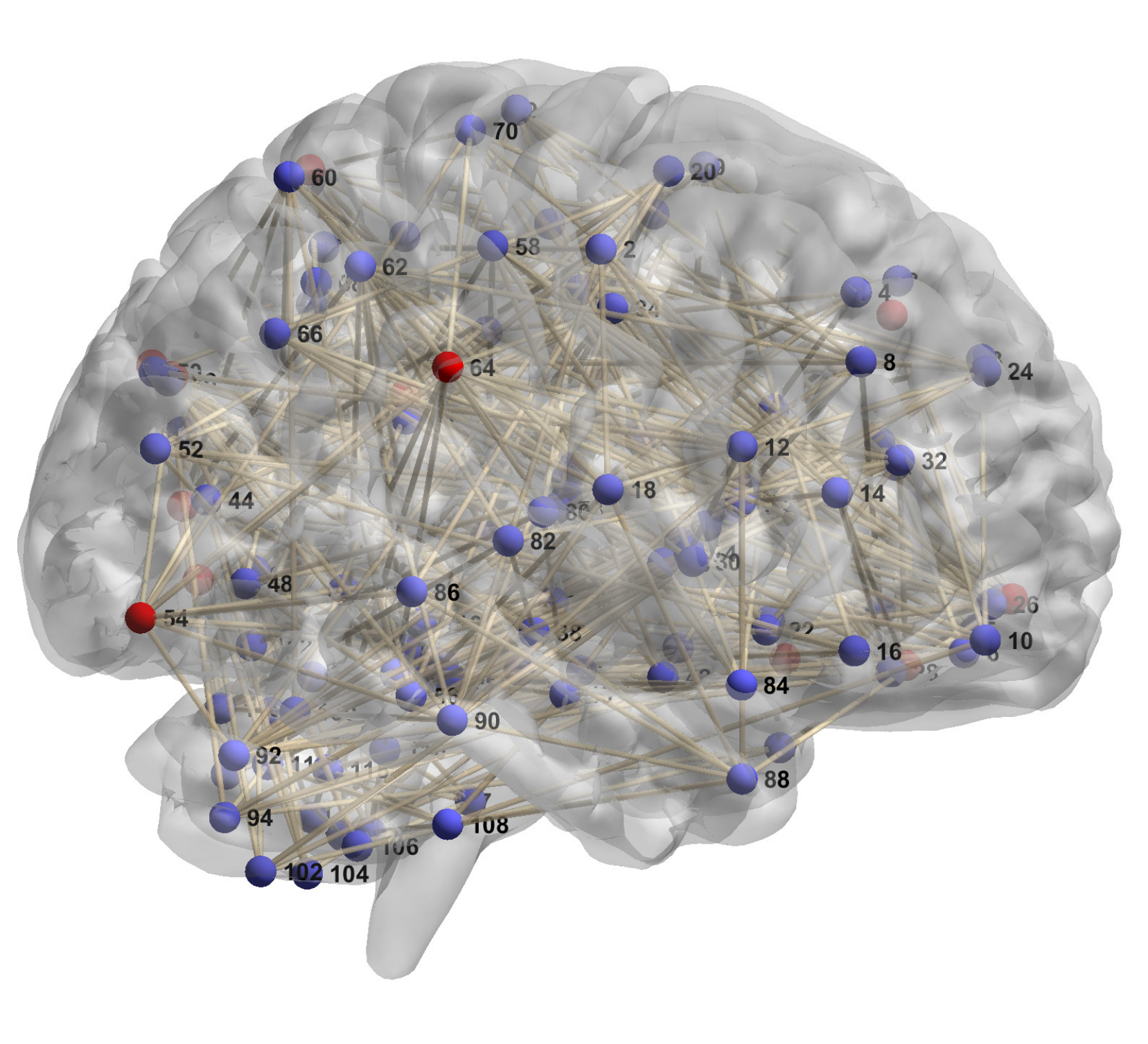}} \
	\subfigure{\includegraphics[scale = 0.29]{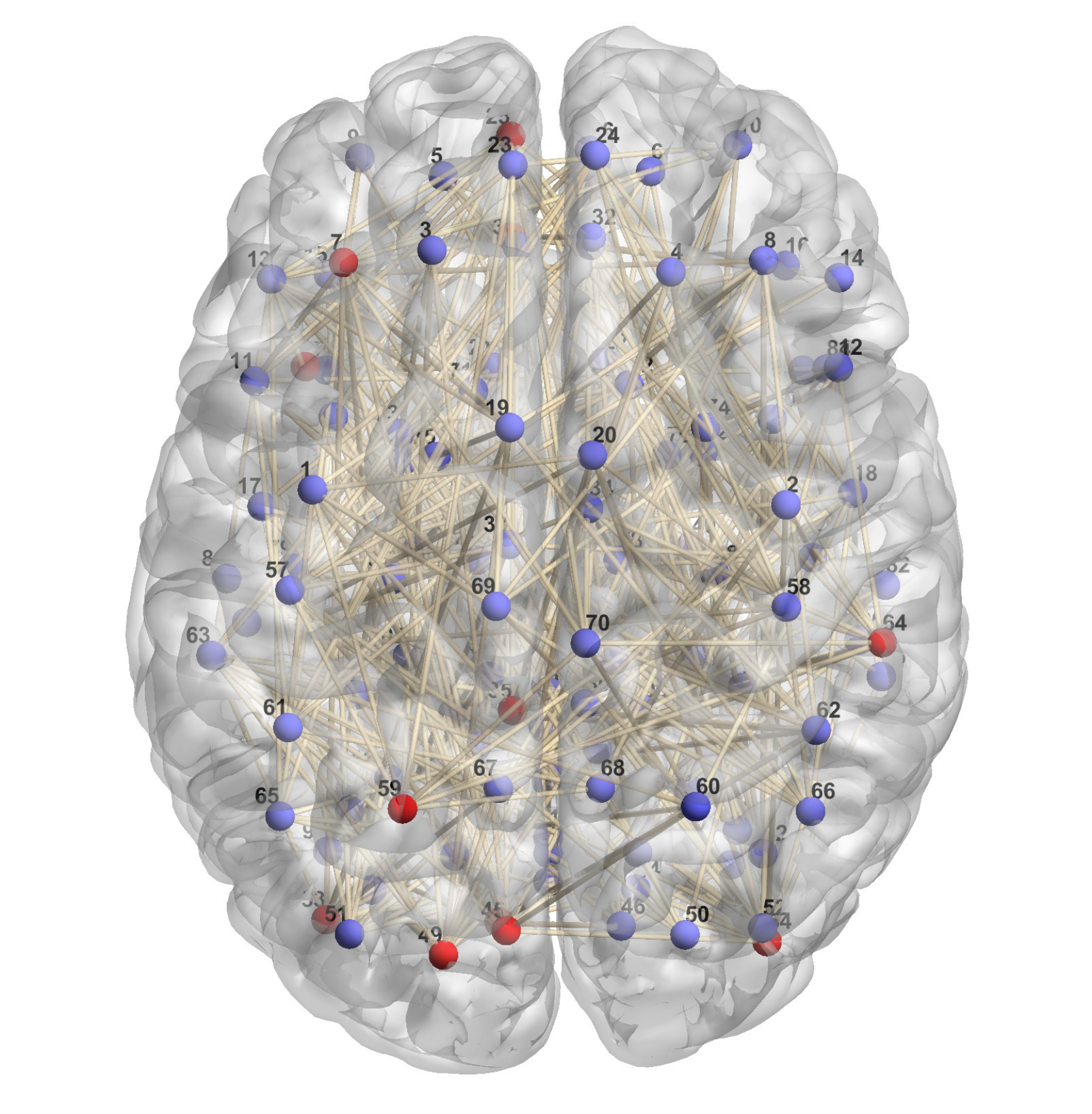}}
	\subfigure{\includegraphics[scale = 0.29]{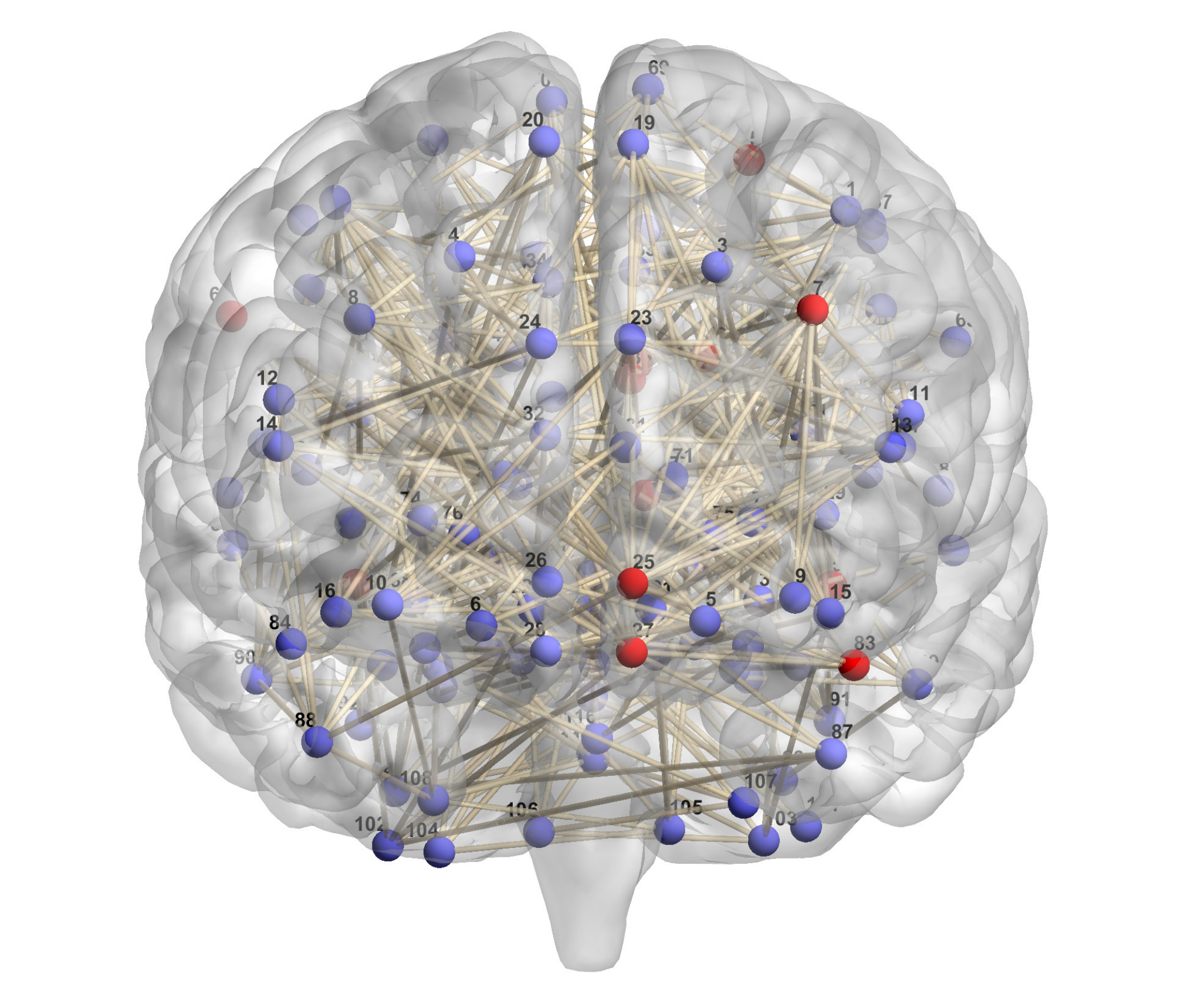}}
	\subfigure{\includegraphics[scale = 0.31]{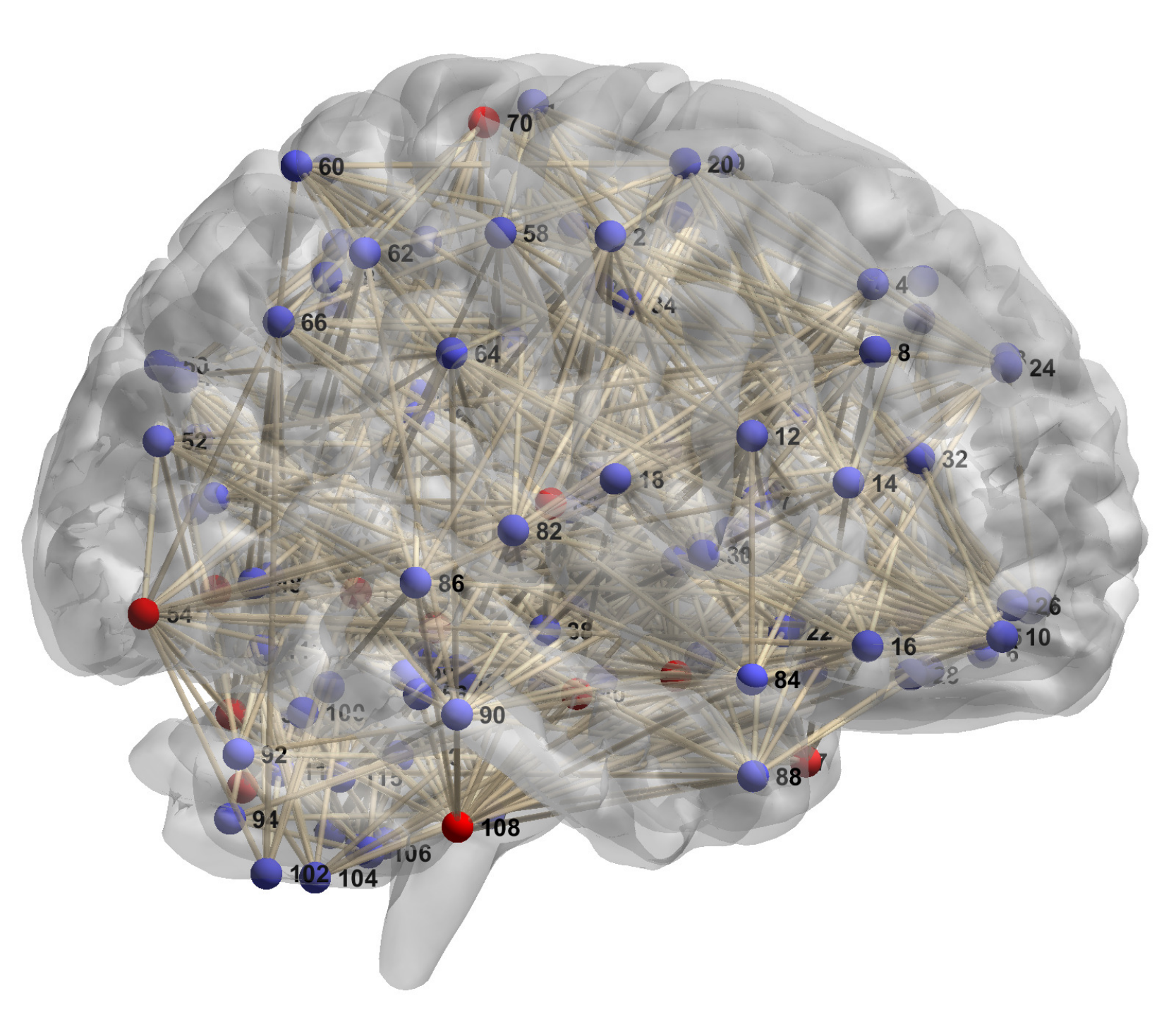}}
	\subfigure{\includegraphics[scale = 0.29]{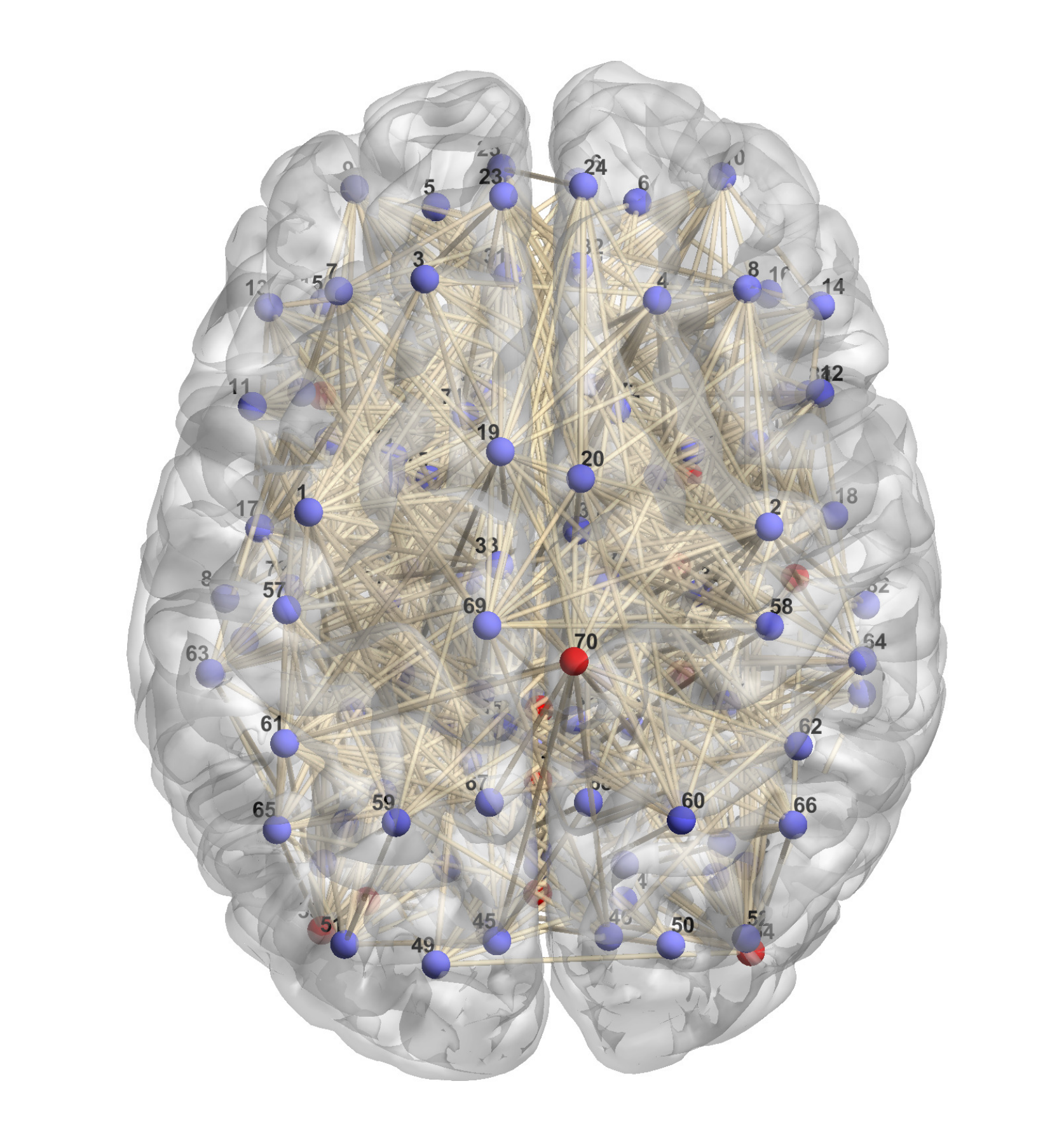}} \
	\subfigure{\includegraphics[scale = 0.29]{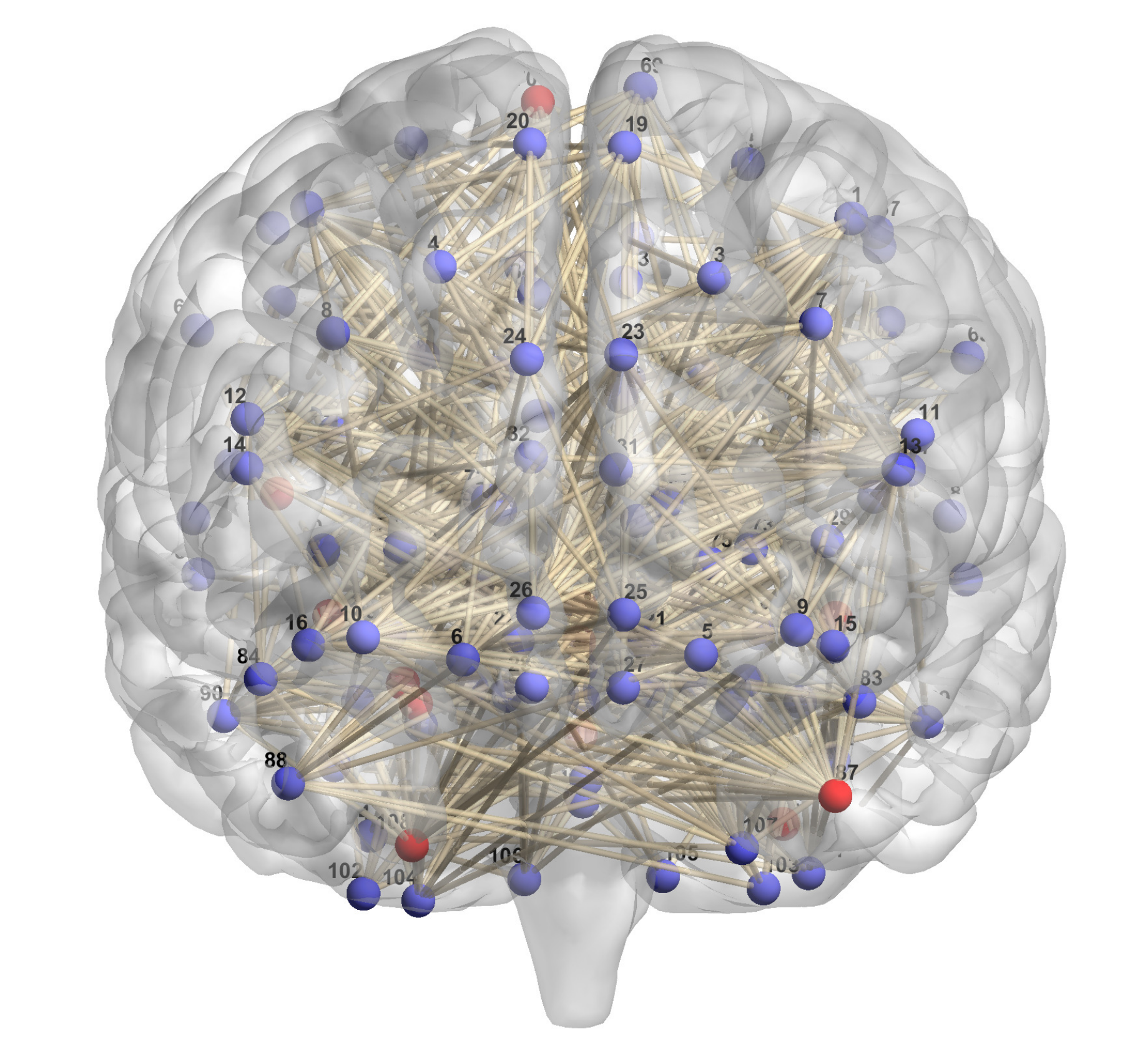}}
	\caption{The differential networks of brain functional connectivity between ADHD and TDC groups. The top rows shows the the edges in the TDC group but not the ADHD group, whereas the bottom rows shows the edges in the ADHD group but not the TDC group. In each row, three different views are also provided: sagittal (left), axial (middle), and coronal (right). Nodes with the top 10\% of degrees are marked by red.}
	\label{diff-brain}
\end{figure}

	As a byproduct of the above procedure, we are also able to reasonably select OHSU, the target site with the lowest relative prediction error in both TDC and ADHD groups, as the target domain to further demonstrate the performance of the proposed transfer learning-based method by conducting more in-depth biomedical exploration. The detected brain networks of TDC and ADHD groups using the proposed method are provided in Figure \ref{abrain}, and it is clear that the two groups are substantially different. To scrutinize their differeces, we plot the differential networks between TDC and ADHD groups in Figure \ref{diff-brain}, with ROIs labeled as the SRI24 code. A cross-reference between the SRI24 code and full names of ROIs can be found in Supporting Information. The top 10\% important hub nodes and their degrees in differential networks are placed in Table \ref{diff.degree} and highlighted in Figure \ref{diff-brain}, many of which have been widely recognized as relevant to ADHD.

	It is evident that the superior frontal gyrus, labeled as 25, has more connections in the TDC group. In fact, it has been found that in the ADHD group, reduced gray matter volumes occurred in this region, and there was a decrease in functional connections between the superior frontal gyrus and other brain regions comparing with the TDC group \citep{zhao2020aberrant}. The functional connectivity mechanism of the inferior occipital gyrus, labeled as 53 and 54, has been recognized significantly different between TDC and ADHD groups, and inattention improvement is related to increased intrinsic brain activity in this region \citep{zhang2020aberrant}. It has been reported that disturbed microstructure of the supramarginal gyrus, labeled as 64, in children with ADHD \citep{griffiths2021structural}. In the detected brain network, the cerebellum inferior, labeled as 108, has more connections in the ADHD group. Actually, the cerebellum has been recognized as an important structure in ADHD pathophysiology, and its abnormalities have been reported in patients with ADHD \citep{stoodley2016cerebellum}. In addition, the decreased cerebellar activation in ADHD has been revealed in many cognitive tasks \citep{valera2010neural}. Moreover, it has been reported that patients with ADHD have a larger probability of activation in the paracentral lobule compared to TDC \citep{dickstein2006neural}, and this region plays an important role in brain functional networks by controling sensory nerves of the contralateral lower limb. In conclusion, the analysis results are basically consistent with the evidence of a large number of neuroscience studies.

	\section{Discussion}
	
	This paper proposes a transfer learning method for tensor GGMs leveraging the separability of its covariance. For each mode, a two-step algorithm is performed to improve the estimation in the target domain by making full use of the information from auxiliary domains, in which we design data-adaptive weights on auxiliary domains that can detect informative auxiliary domains and free from the interference of non-informative auxiliary domains. Theoretically, it has been shown that the estimation error of the proposed transfer learning method can be improved with the increasing sample size from informative auxiliary domains. The condition required for the recovery of graph structures has been relaxed in terms of variable selection. Numerical simulations have been performed to verify the statistical theory and to demonstrate the dominant advantages of the proposed method. The conclusions of real data analysis are also consistent with the existing biological knowledge.
	
	This work has some potential extensions. The development of semi-parametric tensor graphical models is an important refinement to address the non-Gaussian property frequently found in biomedical tensor data. Moreover, it is also worthwhile to explore the tensor-valued differential network model to perform inferential analysis on different edges between two networks, which can replace the current descriptive contrastive patterns between two groups in the ADHD brain network analysis.

\subsection*{Acknowledgment}
This research is supported in part by HK RGC grants GRF-11304520, GRF-11301521, and GRF-11311022.


\bibliographystyle{ECA}
\bibliography{reference}

\clearpage	

\clearpage
\begin{table}[!htb]
	\centering
	\caption{Summary of test errors for each site as the target domain and their sample sizes.}
	\medspace
	\scalebox{0.7}{
		\begin{tabular}{cccccccccc}
			\hline
			&		&	&	\multicolumn{7}{c}{The target domain}	 \\
			\cmidrule(lr){4-10}
			&		&		&	KKI	&	NeuroIMAGE	&	Peking	&	Pittsburgh	&	NYU	&	\textbf{OHSU}	&	WashU	\\
			\hline										
			TDC	&	\multicolumn{2}{c}{sample size}		&	58	&	22	&	114	&	66	&	91	&	40	&	37	\\
			\cdashline{2-10}
			&	absolute error	&	Tlasso	&	0.6773 	&	0.3045 	&	0.4871 	&	0.3914 	&	0.0122 	&	0.4960 	&	0.3962 	\\
			&		&	proposed	&	0.5816 	&	0.2913 	&	0.4506 	&	0.3568 	&	0.0144 	&	0.4179 	&	0.3428 	\\
			&		&	proposed.v	&	0.5829 	&	0.2910 	&	0.4498 	&	0.3514 	&	0.0112 	&	0.4074 	&	0.2939 	\\
			\cdashline{2-10}
			&	relative error	&	proposed	&	0.8588 	&	0.9566 	&	0.9250 	&	0.9115 	&	1.1855 	&	0.8426 	&	0.8652 	\\
			&		&	proposed.v	&	0.8607 	&	0.9557 	&	0.9235 	&	0.8978 	&	0.9169 	&	\textbf{0.8215} 	&	0.7417 	\\
			\hline
			ADHD	&	\multicolumn{2}{c}{sample size}	&	20	&	17	&	90	&	0	&	96	&	30	&	0	\\
			\cdashline{2-10}
			&	absolute error	&	Tlasso	&	0.5002 	&	0.3627 	&	0.6007 	&	-	&	0.0139 	&	0.3503 	&	-	\\
			&		&	proposed	&	0.4461 	&	0.3415 	&	0.5754 	&	-	&	0.0166 	&	0.3072 	&	-	\\
			&		&	proposed.v	&	0.4465 	&	0.3423 	&	0.5754 	&	-	&	0.0165 	&	0.3059 	&	-	\\
			\cdashline{2-10}
			&	relative error	&	proposed	&	0.8919 	&	0.9417 	&	0.9579 	&	-	&	1.1974 	&	0.8770 	&	-	\\
			&		&	proposed.v	&	0.8927 	&	0.9438 	&	0.9579 	&	-	&	1.1907 	&	\textbf{0.8734} 	&	-	\\
			\hline
	\end{tabular}}
	\label{test}
\end{table}

\begin{table}[!htb]
	\centering
	\caption{The top 10\% important hub nodes and their degrees in differential networks of brain functional connectivity between ADHD and TDC groups.}
	\medspace
	\scalebox{0.7}{
		\begin{tabular}{cccc}
			\hline
			& SRI24 code & Full name & Degree	 \\
			\hline	
			TDC-ADHD	&	25	&	Superior frontal gyrus, medial orbital	&	20	\\
			&	27	&	Gyrus rectus	&	16	\\
			&	43	&	Calcarine fissure and surrounding cortex	&	15	\\
			&	45	&	Cuneus	&	15	\\
			&	49	&	Superior occipital gyrus	&	15	\\
			&	35	&	Posterior cingulate gyrus	&	14	\\
			&	53	&	Inferior occipital gyrus	&	14	\\
			&	54	&	Inferior occipital gyrus	&	14	\\
			&	59	&	Superior parietal gyrus	&	13	\\
			&	64	&	Supramarginal gyrus	&	13	\\
			&	83	&	Temporal pole: superior temporal gyrus	&	13	\\
			&	7	&	Middle frontal gyrus	&	12	\\
			\hline	
			ADHD-TDC	&	87	&	Temporal pole: middle temporal gyrus	&	29	\\
			&	108	&	Cerebellum Inferior	&	26	\\
			&	113	&	Vermis	&	25	\\
			&	54	&	Inferior occipital gyrus	&	23	\\
			&	80	&	Heschl gyrus	&	23	\\
			&	93	&	Cerebellum Inferior	&	23	\\
			&	40	&	Parahippocampal gyrus	&	21	\\
			&	110	&	Vermis	&	21	\\
			&	111	&	Vermis	&	21	\\
			&	42	&	Amygdala	&	20	\\
			&	53	&	Inferior occipital gyrus	&	20	\\
			&	70	&	Paracentral lobule	&	20	\\
			\hline
	\end{tabular}}\\
\footnotesize{$^*$TDC-ADHD: the differential network consisting of the edges in the TDC group but not the ADHD group; ADHD-TDC: the differential network consisting of the edges in the ADHD group but not the TDC group.}
	\label{diff.degree}
\end{table}

\end{document}